%% file: main.tex
\documentclass[10pt,journal,compsoc]{IEEEtran}
\usepackage[utf8]{inputenc} 
\usepackage[T1]{fontenc}  
\usepackage[breaklinks=true,
            colorlinks,
            linkcolor = red,
            urlcolor  = purple, 
            citecolor = blue,
            bookmarks = black]{hyperref}  
\usepackage{amsmath,amsfonts}
\usepackage{bbding}
\usepackage{utfsym}
\usepackage{algorithmic}
\usepackage{algorithm}
\usepackage{array}
\usepackage[caption=false,
font=normalsize,
labelfont=sf,
textfont=sf]{subfig}
\ifCLASSOPTIONcompsoc
\usepackage[nocompress]{cite}
\else
\usepackage{cite}
\fi

\ifCLASSINFOpdf
\usepackage{textcomp}
\usepackage{stfloats}
\usepackage{url}
\usepackage{verbatim}
\usepackage{graphicx}
\usepackage{cite}
\usepackage{multirow}
\usepackage{booktabs} 
\renewcommand{\paragraph}[1]{\vspace{.25em}\noindent\textbf{#1.}}

\begin{document}
\title{GMTalker: Gaussian Mixture-based Audio-Driven Emotional Talking Video Portraits}

\author{Yibo Xia$^*$, 
        Lizhen Wang, 
        Xiang Deng, 
        Xiaoyan Luo~\IEEEmembership{Member,~IEEE,} 
        Yunhong Wang~\IEEEmembership{Fellow,~IEEE,}
        and Yebin Liu~\IEEEmembership{Member,~IEEE}
\IEEEcompsocitemizethanks{
\IEEEcompsocthanksitem $^*$ Work done during an internship at Tsinghua University.
\IEEEcompsocthanksitem Yibo Xia, Xiaoyan Luo, and Yunhong Wang are with Beihang University, Beijing, 100191, P.R. China.
\IEEEcompsocthanksitem Lizhen Wang, Xiang Deng, and Yebin Liu are with Tsinghua University, Beijing 100084, P.R. China.
\IEEEcompsocthanksitem Corresponding author: Xiaoyan Luo.
}
}

\markboth{Journal of \LaTeX\ Class Files,~Vol.~14, No.~8, August~2021}%
{Shell \MakeLowercase{\textit{et al.}}: A Sample Article Using IEEEtran.cls for IEEE Journals}


\IEEEtitleabstractindextext{
\begin{abstract}
Synthesizing high-fidelity and emotion-controllable talking video portraits, with audio-lip sync, vivid expressions, realistic head poses, and eye blinks, has been an important and challenging task in recent years. Most existing methods suffer in achieving personalized and precise emotion control, smooth transitions between different emotion states, and the generation of diverse motions. To tackle these challenges, we present GMTalker, a Gaussian mixture-based emotional talking portraits generation framework.
Specifically, we propose a Gaussian mixture-based expression generator that can construct a continuous and disentangled latent space, achieving more flexible emotion manipulation. Furthermore, we introduce a normalizing flow-based motion generator pretrained on a large dataset with a wide-range motion to generate diverse head poses, blinks, and eyeball movements.
Finally, we propose a personalized emotion-guided head generator with an emotion mapping network that can synthesize high-fidelity and faithful emotional video portraits.
Both quantitative and qualitative experiments demonstrate our method outperforms previous methods in image quality, photo-realism, emotion accuracy, and motion diversity.
\end{abstract}
\begin{IEEEkeywords}
Facial Animation, Gaussian Mixture Model, Talking Video Portrait, Continuously Emotion Manipulation
\end{IEEEkeywords}
}
\maketitle
\IEEEdisplaynontitleabstractindextext
\IEEEpeerreviewmaketitle

\input{sec/1_intro}

\input{sec/2_related_work}

\input{sec/3_methods}

\input{sec/4_experiment}
\input{sec/5_conclusion}

\bibliographystyle{IEEEtran}
\bibliography{main}


\vfill
\clearpage

\appendices
\input{sec/suppl_1}

\end{document}

%% file: sec/1_intro.tex
\IEEEraisesectionheading{\section{Introduction}\label{sec:intro}}

\begin{figure*}
  \centering
  \includegraphics[width=\textwidth]{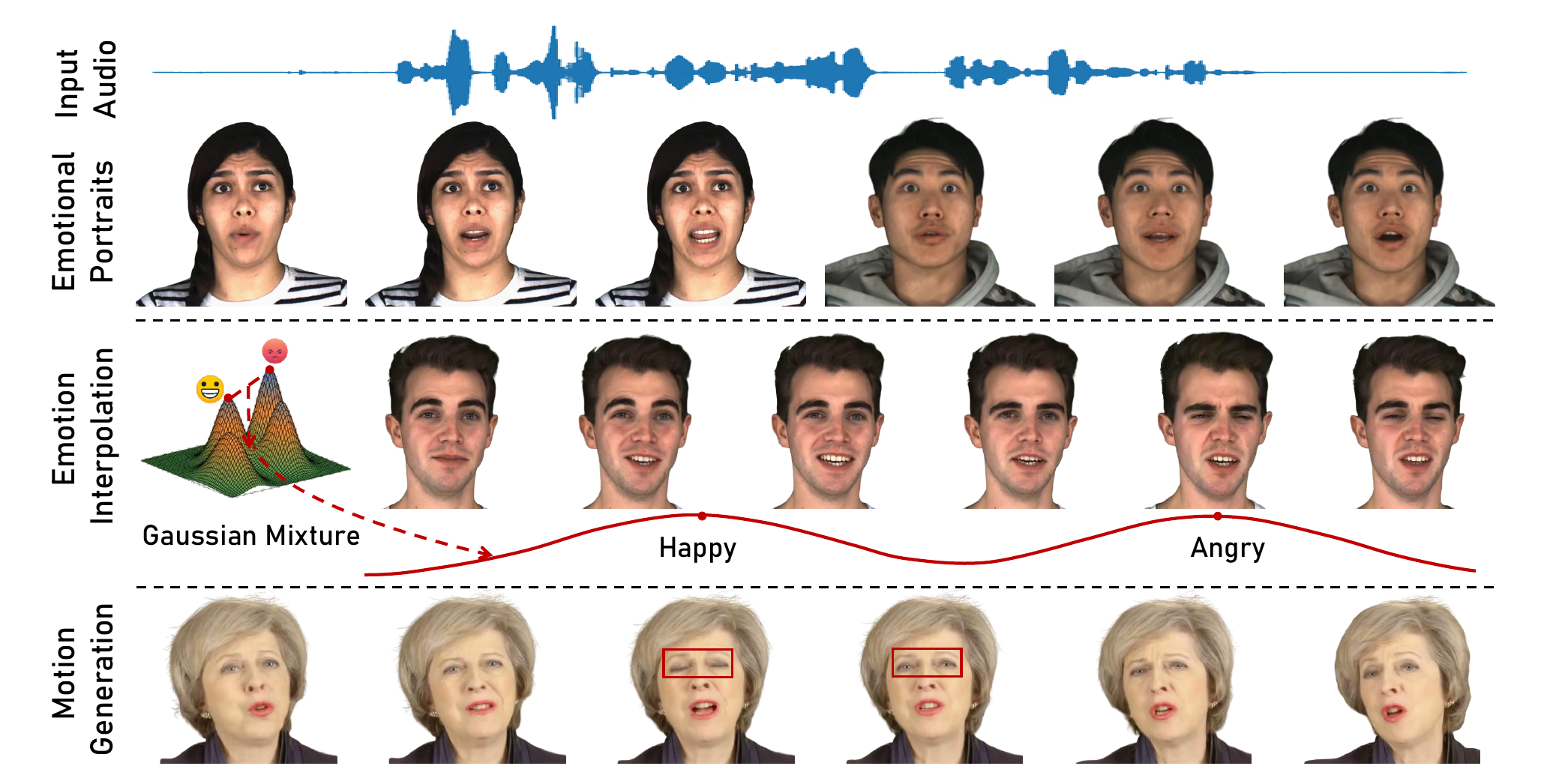}
  \vspace{-0.8cm}
  \caption{GMTalker. Given the driving speech and emotion label, our method can generate high-fidelity and faithful emotional talking video portraits with diverse motions. Emotions can be freely manipulated within our continuous and disentangled Gaussian mixture distributed latent space. Additionally, our method can also predict motions from the input speech, including head poses, eye blinks, and gaze.}
  \label{fig:teaser}
\vspace{-0.4cm}
\end{figure*}


\IEEEPARstart{A}{udio-driven} talking video portraits have drawn much research interest due to their broad applications in education, filmmaking, virtual digital human and entertainment industry, etc. It aims to produce audio-lip sync, photo-realistic, freely controllable video portraits given a driven speech. Actually, facial emotions and motions in other aspects, including head pose, eye blinks, and gaze, play a crucial role in generating photo-realistic and vivid video portraits. However, existing methods encounter challenges in achieving accurate and continuous emotional control, along with generating diverse motions and personalized speaking styles.

Previous methods~\cite{prajwal2020lip,sun2022masked,cheng2022videoretalking,guan2023stylesync} focus on audio-lip synchronization across different speakers, ignoring the control of facial emotion and motion generation. 
Some works pay attention to emotion control by either learning emotions from audio~\cite{ji2021audio,tan2023emmn,xu2023high} or adding emotional source videos~\cite{ji2022eamm,liang2022expressive,wang2023progressive,ma2023styletalk}, which will introduce ambiguities.
More recent methods~\cite{eskimez2021speech,wang2020mead,sinha2022emotion,gururani2023space,gan2023efficient} focus on generating emotional expressions that are consistent with the desired emotion label, providing a more reasonable and controllable approach for synthesizing emotional talking video portraits. However, they still face challenges in achieving precise emotion control or continuously interpolating between different emotion states.
This limitation stems from their approach of conditioning the emotion label into an emotion-agnostic framework through a one-hot vector (representing a discrete emotion space)~\cite{eskimez2021speech,wang2020mead,sinha2022emotion, gururani2023space} or deep emotional prompts (representing an entangled emotion space)~\cite{gan2023efficient} to implicitly learn the mapping between emotion and facial expression.  
The majority of these approaches fail to model a continuous and disentangled emotion space for better interpolation properties as well as more precise emotion control.
To address this problem, we propose a Gaussian Mixture based Expression Generator (GMEG) which explicitly learns a conditional Gaussian mixture distribution among audio, emotion, and 3DMM expression coefficients. 
Our insight is to construct a continuous and disentangled latent space, where each Gaussian component represents specific emotional properties of the data, and these components are highly decoupled from each other. By leveraging this learned Gaussian mixture latent space, we achieve precise emotion control and smooth emotional transition.

Besides emotional facial expressions, the motion of other factors, such as head pose, eye blink, and gaze, are essential to the realism of synthesized videos.
Some research~\cite{chen2020talking,yi2020audio,zhou2021pose,lu2021live,zhang2021facial,wang2021audio2head,wang2022one,liu2023moda,yu2023talking} model the correspondence between audio and motion. However, they have not considered emotional control, and only a few works~\cite{gururani2023space,tan2023emmn} can control both emotion and motions simultaneously. 
Moreover, they encounter a so-called ``mean motion'' challenge, which means the synthesized motion tends to be over-smoothing and lacks diversity.
To overcome their weakness, we propose a Normalizing Flow-based Motion Generator (NFMG), which enhances the motion prior by incorporating normalizing flow and learns the one-to-many mapping between the speech and motions. 
Moreover, to fully leverage the normalizing flow's potential for fitting complex data distributions, we pre-train proposed NFMG on Vox2celeb2~\cite{chung2018voxceleb2} dataset with wide-range head and eye movements. In this way, we can obtain a more diverse motion prior and alleviate the ``mean motion'' issue. Additionally, our method can independently control expressions and motions by utilizing a parametric facial model~\cite{wang2022faceverse}. 

Finally, existing methods~\cite{ji2022eamm,gan2023efficient,gururani2023space} lack consideration for generating emotional portraits faithful to a specific person. To address this limitation, we employ the stylized head generator, StyleUNet~\cite{wang2023styleavatar}, which reconstructs the personalized style of a target identity using a latent code. However, owing to the highly coupled latent space of StyleUNet, it struggles to explicitly control the desired emotion. Therefore, we introduce an Emotion Mapping Network (EMN) to branch each emotion mode corresponding to the available sub-domain, which can control detailed emotion-related styles of the target person. 
Consequently, we can synthesize emotional portraits with personalized speaking styles.

We conduct comprehensive emotion interpolation comparison experiments, evaluating the smoothness and precision of the emotion transition from both quantitative and qualitative perspectives. 
Compared with other state-of-the-art methods, the proposed GMTalker shows smoother and more precise emotion transitions while maintaining accurate lip synchronization.
Our main contributions can be summarized as follows:

\begin{itemize}
\item We propose a Gaussian mixture-based expression generator (GMEG) to disentangle different emotion states in continuous latent space, thereby achieving more precise emotion control and better interpolation properties.

\item We present a normalizing flow-based motion generator (NFMG) pretrained on a large dataset with wide-range motions to generate diverse motion coefficients, including head poses, eye blinks, and gaze.

\item We introduce a personalized emotion-guided head generator with an emotion mapping network (EMN) that synthesizes high-fidelity and faithful emotional video portraits with personalized speaking styles.

\end{itemize}

%% file: sec/2_related_work.tex
\vspace{-0.2cm}
\section{Related Work}
\label{sec:related_work}

Most existing works can be roughly classified into digital 3D human faces~\cite{karras2017audio,richard2021meshtalk,fan2022faceformer,xing2023codetalker,thambiraja2023imitator,peng2023emotalk,danvevcek2023emotional} and realistic human portraits~\cite{suwajanakorn2017synthesizing,prajwal2020lip,chen2019hierarchical,zhou2020makelttalk}, according to their output. The methods that animate 3D models of faces map the input speech to 3D mesh via carefully designed architecture. However, their applicability is limited due to the requirement for expensive 3D training data. Thus, we focus on generating photo-realistic talking video portraits.

\paragraph{Speech-Driven Talking Video Portraits} 
Most of the existing methods pay attention to generating movements in the mouth region~\cite{sadoughi2019speech,kr2019towards,prajwal2020lip,park2022synctalkface,sun2022masked,cheng2022videoretalking,guan2023stylesync,ki2023stylelipsync,zhang2023dinet,wu2023speech2lip, wang2023seeing, shen2023difftalk}. For instance, Wav2Lip~\cite{prajwal2020lip} inpaints the lower half of the face using an expert SyncNet~\cite{chung2017out} to align the speech and mouth region.
These lines of research, leaving the remaining video stationary, can only synthesize the lower half face. 
Other methods aim to generate the whole face by either wrapping the reference image according to input speech~\cite{zhang2021flow,wang2021audio2head,wang2022one} or extracting face and audio features as a fused input to the decoder model~\cite{zhou2019talking,zhou2021pose,sun2021speech2talking,wang2023lipformer}. However, directly modeling the correspondence between audio and dynamic facial expressions in an end-to-end manner struggles to control head motion and synthesize high-quality face images.

Recently, with the development of 3D face reconstruction~\cite{deng2019accurate,wang2022faceverse}, some works leverage explicit 2D/3D facial landmarks~\cite{suwajanakorn2017synthesizing,chen2019hierarchical,zhou2020makelttalk,das2020speech,lu2021live,liu2023moda,zhong2023identity} or 3D face models ~\cite{yi2020audio,thies2020neural,chen2020talking,zhang2021facial,song2022everybody,ma2023styletalk,zhang2023sadtalker} to reconstruct interpretable landmarks or face parameters, and then translate them to photo-realistic results.
MakeItTalk~\cite{zhou2020makelttalk} utilizes disentangled speech content and speaker identity features to animate the facial landmarks of a provided portrait. 
Benefiting from the controllability of 2D/3D representations, some methods\cite{chen2020talking,yi2020audio,lu2021live,zhang2021facial,zhang2023sadtalker,liu2023moda} achieve explicitly motion control, which makes the synthesized portraits more realistic. 
LSP~\cite{lu2021live} leverages a probabilistic autoregressive module to reconstruct dynamic landmarks. 
EMO~\cite{tian2024emo} leverages the power of the diffusion model to directly generate video portraits with the controllable head motion, achieving excellent performance. However, none of them has considered emotional control which is a key factor in generating realistic portraits.

\paragraph{Emotional Talking Video Portraits} 
Emotion significantly impacts the realism of synthesized portraits.
Recently, some works have paid attention to controlling the emotion of the output portraits. 
EAMM~\cite{ji2022eamm}, GC-AVT~\cite{liang2022expressive}, Styletalk~\cite{ma2023styletalk}, and PD-FGC~\cite{wang2023progressive} control emotion by external emotional source videos, inevitably introducing a semantic leakage problem. EVP~\cite{ji2021audio}, EMMN~\cite{tan2023emmn} directly identify emotion from labeled audio. However, determining emotions from input audio only may introduce ambiguities~\cite{ji2022eamm}. 
Other works ETK~\cite{eskimez2021speech}, MEAD~\cite{wang2020mead}, Sinha \textit{et al}~\cite{sinha2022emotion}, SPACE~\cite{gururani2023space}, EAT~\cite{gan2023efficient} learn the inherent correspondence among emotion, audio, and facial expression implicitly via conditioning the emotion-agnostic network with emotion labels. However, none of them explicitly models a continuous and disentangled emotion space, leading to poor emotion interpolation properties and inaccurate emotion generation. Our method animates the emotion-controllable talking video portraits, including facial expression, head pose, blinks, and eye gaze.

%% file: sec/3_methods.tex
\vspace{-0.3cm}
\section{Method} \label{sec:methods}

\begin{figure*}
  \centering
  \includegraphics[width=\textwidth]{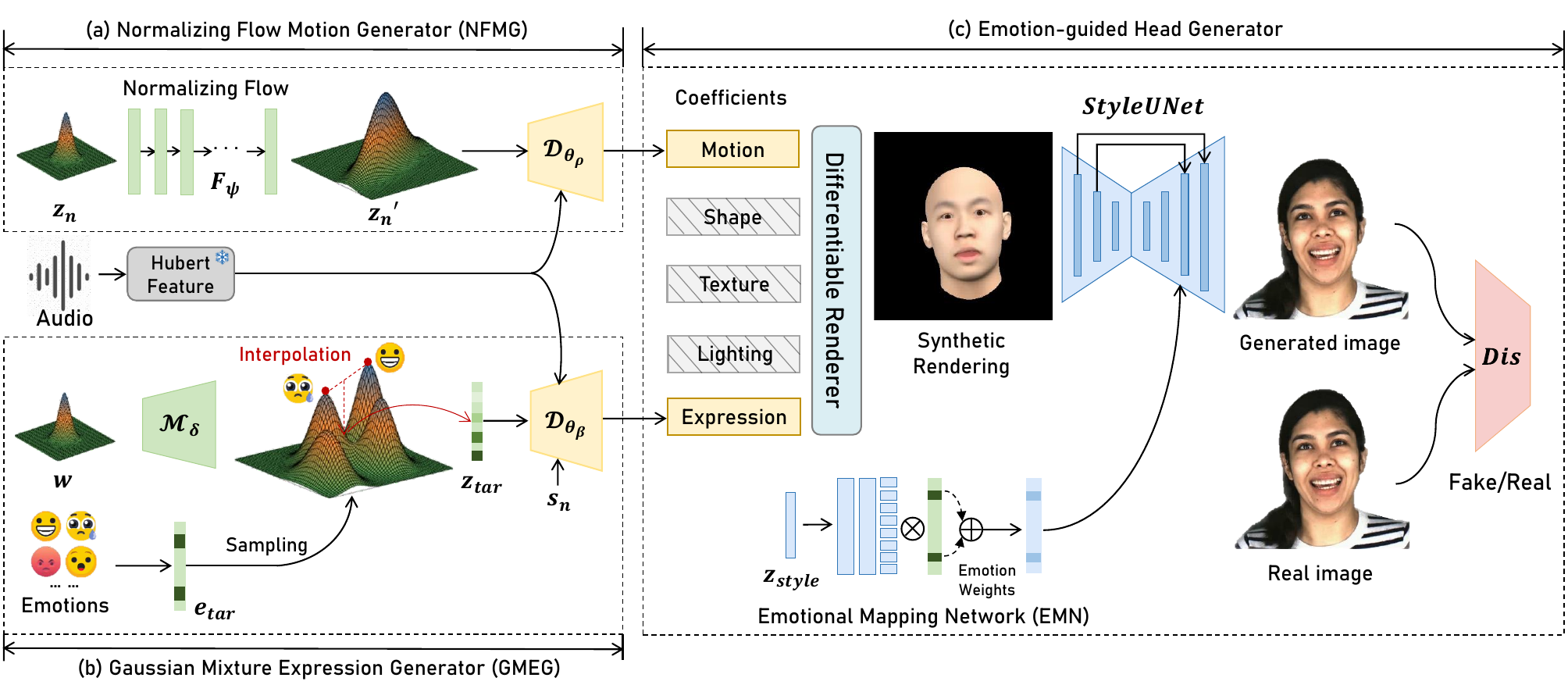}
  \vspace{-0.6cm}
  \caption{Pipeline of GMTalker. Our framework consists of three parts: 
  (a) In Section~\ref{GMVAE}, given the input speech and emotion weights label, we propose GMEG to generate 3DMM expression coefficients sampling from Gaussian mixture latent space. 
  (b) In Section~\ref{FVAE}, we introduce NFMG to predict motion coefficients from the audio, including poses, eye blinks, and gaze. 
  (c) In Section~\ref{StyleUNet}, we render these coefficients to 3DMM renderings for the target person and then use an emotion-guided head generator with EMN to synthesize photo-realistic video portraits with personalized style.}
  \label{fig:pipeline}
\vspace{-0.4cm}
\end{figure*}

We present GMTalker, a Gaussian mixture-based audio-driven emotional video portrait generation framework taking the 3DMM as the intermediate representation.
Given an audio and an emotion label as input, our system produces a talking video of a target person. It includes two generators for emotional expression coefficients and motion-related coefficients, as well as an emotion-guided head generator. The whole pipeline of our proposed method is illustrated in Fig.~\ref{fig:pipeline}.
Specifically, we first extract per-frame 3DMM expression and motion coefficients by fitting a face parametric model (Section~\ref{Preliminaries}). 
Then, we propose a Gaussian mixture expression generator (GMEG) in Section~\ref{GMVAE} to generate emotional 3DMM expression coefficients from an audio and emotion label by learning a Gaussian mixture latent space.
Meanwhile, we present a normalizing flow-based motion generator (NFMG) in Section~\ref{FVAE} to produce diverse head poses, gazes, and eye blink coefficients. 
Finally, we present an emotion-guided head generator with an Emotion Mapping Network (EMN) in Section~\ref{StyleUNet} to generate photo-realistic emotional portraits with personalized speaking styles from the generated expression and motion coefficients. 

\vspace{-0.4cm}
\subsection{3D Head Representation and Data Preprocessing} \label{Preliminaries}

Given a $T$-frame emotional monocular portrait video of the target person, we first perform parametric model fitting to extract 3DMM coefficients as our intermediate representation and generate 3DMM renderings for training. We utilize FaceVerse~\cite{wang2022faceverse} for the following reasons. First, the expression and shape bases of FaceVerse are rich and diverse, enabling it to effectively capture complex and emotion-related expressions across different identities compared with other facial models~\cite{paysan20093d,li2017learning,deng2019accurate}. Second, it excels in tracking stable head pose and capturing eyeballs and eye blinks.
The 3D face shape of FaceVerse $S$ can be formulated as:
\begin{equation}
    S=\overline{S}+\gamma B_{shape} + \beta B_{exp},
\end{equation}
where $\overline{S}$ is the mean shape, $B_{shape}$ and $B_{exp}$ are the bases of shape and expression. Expression
coefficients $\beta\in\mathbb{R}^{169}$, blink coefficients $\theta_{blink}\in\mathbb{R}^2$, translation $t\in\mathbb{R}^3$, scale $s$, and the rotations of the head and two eyeballs $r_1\in\mathbb{R}^3$, $r_2\in\mathbb{R}^2$, $r_3\in\mathbb{R}^2$ are optimized using differentiable renderer in~\cite{wang2023styleavatar} as a frame by frame manner.
To generate identity-irrelevant coefficients, we only optimize shape coefficients $\gamma\in\mathbb{R}^{150}$ in the first frame for the target speaker following~\cite{zhang2023sadtalker,ren2021pirenderer,wang2023styleavatar}.
Finally, we obtain a sequence of facial expression coefficients, $\{\beta\}_{t=1}^T$, that is rich in emotional information, and a sequence of motion coefficients, $\{\rho\}_{t=1}^T=\{[r_1,t], r_2, r_3, \theta_{blink}\}_{t=1}^T$, which are capable of representing realistic movements.
In terms of audio processing, we employ a pretrained HuBERT model~\cite{hsu2021hubert} to extract the audio feature $\{a\}_{t=1}^T$, following methods~\cite{ye2023geneface,ye2024real3d}.


\vspace{-0.2cm}
\subsection{Gaussian Mixture Expression Generator} \label{GMVAE}
Input with the audio and an emotion weight label, we propose a Transformer-based GMEG to generate emotional expression coefficients. We consider the audio-driven emotional expression generation as a conditional generation task and explicitly model the conditional distribution between the input audio feature $\{a\}_{t=1}^T$, emotion weight label $e$, and facial expression coefficients $\{\beta\}_{t=1}^T$. 

Since previous methods struggle to model a continuous and disentangled emotion space with better interpolation properties and more precise emotion control, we utilize a Gaussian mixture distribution to model the emotion latent space for expression generation, inspired by GMVAE~\cite{sohn2015learning}.
Our insight is that the observed emotion data follows a Gaussian mixture distribution, and we restrict the distribution of the latent code to their corresponding emotion modes. 
This approach allows us to model a continuous and disentangled Gaussian mixture latent space, in which we can easily control emotion and smoothly interpolate between different emotion states. 

\subsubsection{Preliminary}

Our GMEG can be mathematically formulated as a joint distribution:
\begin{equation}
p_{\beta,\theta}(\beta,{z},{w},{e},{a})=p({w})p({e})p_{\delta}({z}|{w},{e})p_{\theta_\beta}({\beta}|{z},{a}), 
\vspace{-0.1cm}
\end{equation}
which means our generative model will generate an observed expression coefficient $\beta$ from the audio $a$, the latent variables $w$, $z$, and the emotion label $e$. Specifically, the latent variable $w$ follows the normal distribution ${w}\sim\mathcal{N}(0,{I})$, $z$ is a latent variable, and the emotion label $e$ follows the uniform distribution ${e}\sim\mathcal{U}(0,{K})$, $p(e=k)={\pi_k}=1/K$, where $K$ is the number of components in the mixture (i.e. the number of emotion types in datasets).

Then, by sample from $w$-space conditioned on various emotion $e_{k}$, we can model a conditioned Gaussian mixture distribution ${z}|{e},{w}$ (i.e. Gaussian mixture latent space):

\begin{equation}
p_{\delta}({z}|{w},{e})=\sum_{k=1}^{K}{\pi_k}\mathcal{N}({z};{\mu}_{\delta}^{k}(w),\Sigma_{\delta}^{k}({w})), 
\vspace{-0.2cm}
\end{equation}
where ${\mu}_{\delta}^k$ and $\Sigma_{\delta}^k$ represent a set of $K$ means and variances of Gaussian mixture modeled by a neural network. 
Finally, we can generate expression coefficients $\beta_{1:t}$ by a decoder with latent variable $z$ and audio $a_{1:t}$ as input.

\subsubsection{Architecture}

We employ a VAE~\cite{kingma2013auto} paradigm to learn the above Gaussian mixture latent space, which includes an encoder, a mixture-of-Gaussian (MoG) mapper, and a decoder. To capture long-range context dependencies and process arbitrary-length sequences, we further employ Transformer architectures~\cite{vaswani2017attention} to model all of these networks and obtain sequence-level embeddings for both expression coefficients and audio features.

\begin{figure*}
  \centering
  \includegraphics[width=0.96\textwidth]{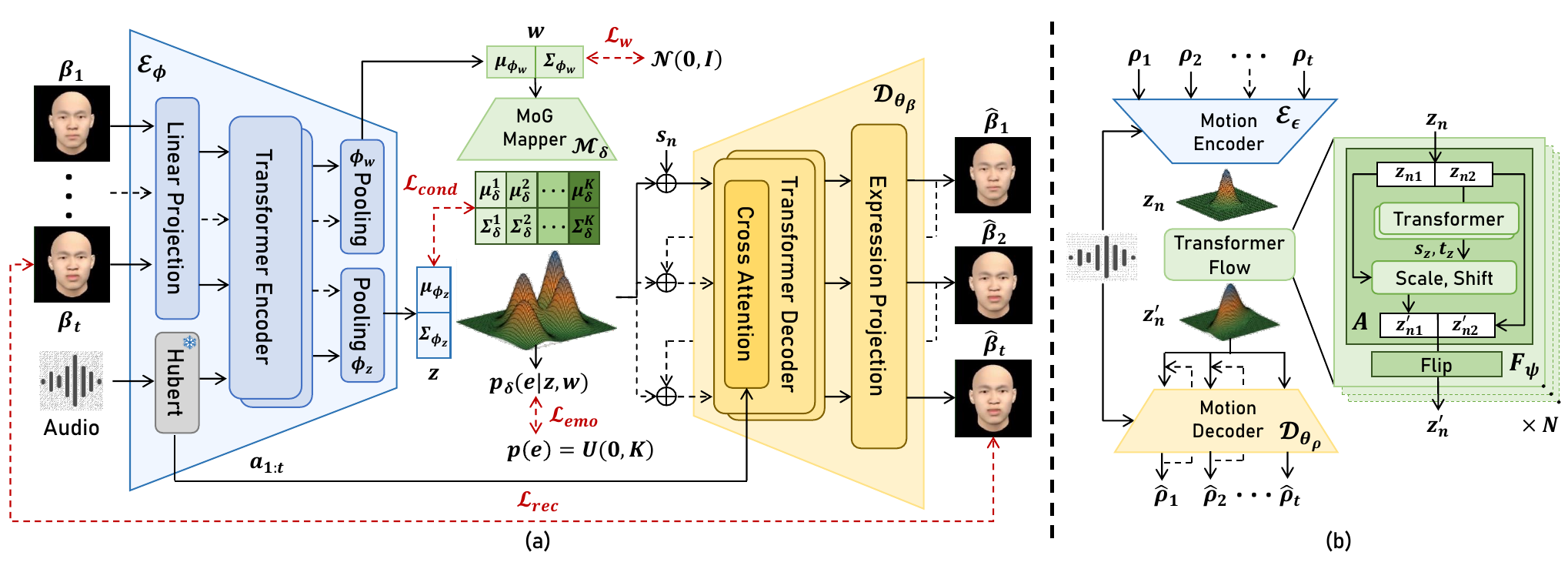}
  \vspace{-0.3cm}
  \caption{The training process of our proposed GMEG and NFMG.
  (a) We autoregressively reconstruct facial expression coefficients $\hat{\beta}_{1:t}$ from input audio $a_{1:t}$ and emotion label $e$ by optimizing four loss: $\mathcal{L}_{rec}$, $\mathcal{L}_{cond}$, $\mathcal{L}_{w}$, $\mathcal{L}_{emo}$. 
  (b) Our NFMG generates diverse motions $\hat{\rho}_{1:t}$ from audio, including head poses, eye blinks, and gaze, by learning a Transformer normalizing flow-based VAE.
  }
  \label{fig:training_pipe}
\vspace{-0.5cm}
\end{figure*}

\noindent\textbf{Encoder.}
As highlighted by the blue part in Fig.~\ref{fig:training_pipe}. (a), our encoder $\mathcal{E}_{\phi}$ aims at learning two approximate posterior distributions: a normal distribution $q_{\phi_z}({z}|\beta, a)=\mathcal{N}(z;\mu_{\phi_z}(\beta, a),\Sigma_{\phi_z}(\beta, a))$ and Gaussian mixture distribution $q_{\phi_w}({w}|\beta, a)=\mathcal{N}(w;\mu_{\phi_w}(\beta, a),\Sigma_{\phi_w}(\beta, a))$, taking expression coefficients $\beta_{1:{t}}$ and audio features $a_{1:t}$ as input:
\begin{equation}
    z, w = \mathcal{E}_{\phi}(\beta_{1:t}, a_{1:t}).
\vspace{-0.1cm}
\end{equation}
We first embed the input expression coefficients into a $d$-dimensional space via a linear projection. Then, we combine these projected features with audio features as the input of sinusoidal positional encoding to provide temporal order information periodically~\cite{vaswani2017attention}. We use 8 Transformer encoder layers to model capture long-range context, followed by two average pooling layers to produce two groups of distribution parameters: $\mu_{\phi_w}$, $\Sigma_{\phi_w}$, $\mu_{\phi_z}$ and $\Sigma_{\phi_z}$.

\noindent\textbf{Mixture-of-Gaussian Mapper.}
To generate the conditioned Gaussian mixture distribution $p(z|w)$ from the latent variable $w$, we propose a Transformer-based MoG mapper $\mathcal{M_\delta}$ parameterized by $\delta$, depicted in the green part of Fig.~\ref{fig:training_pipe}. (a). 
It consists of 8 Transformer encoder layers, without positional encodings.
Our MoG mapper outputs a set of $K$ means ${\mu}_{\delta}^k$ and variances ${\Sigma}_{\delta}^k$ of Guassian mixture, written as:
\begin{equation}
\vspace{-0.1cm}
    {\mu}_{\delta}^k, {\Sigma}_{\delta}^k = \mathcal{M_\delta}(w).
\end{equation}

\noindent\textbf{Decoder.}
Given the latent variable $z$, audio feature $a_{1:t}$, and the personalized learnable embedding $s_{n}$ of speaker $n$, we autoregressively generate emotional expression coefficients $\beta_{1:t}$ by a Transformer-based decoder parametrized by $\theta_{\beta}$,
\begin{equation}
\vspace{-0.1cm}
    \hat{\beta}_{t} = \mathcal{D}_{\theta_{\beta}}(z, a_{1:t}, \hat{\beta}_{1:t-1}, s_n).
\end{equation}
Our decoder $\mathcal{D}_{\theta_{\beta}}$ consists of 8 Transformer decoder layers with a linear expression projection layer, shown in the yellow part of Fig.~\ref{fig:training_pipe}. (a). We add a latent variable $z$ to a sequence of positional encodings as the Transformer input embeddings and then predict the current expression coefficient $\hat{\beta}_{t}$ conditioned on the previous expression coefficients $\hat{\beta}_{1:t-1}$.
The personalized learnable embedding $s_{n}$ can be regarded as the initialized expression coefficients, serving as a starting guide.
Notably, we enhance the Transformer decoder with causal self-attention to capture dependencies within the past expression coefficient sequence, and with cross-modal attention to align the audio and expressions, inspired by~\cite{fan2022faceformer}.

\subsubsection{Training Loss} 
Due to the introduction of the Gaussian mixture prior, the optimization objective of VAE has some changes and no longer offers a simple analytical solution. 
Therefore, we optimize our GMEG using the log-evidence lower bound (ELBO) loss following~\cite{dilokthanakul2016deep}, which can be written as:
\begin{equation}
\vspace{-0.1cm}
\mathcal{L}_{exp}=\lambda_{rec}\mathcal{L}_{rec}+\lambda_{cond}\mathcal{L}_{cond}+\lambda_{w}\mathcal{L}_{w}+\lambda_{emo}\mathcal{L}_{emo}.
\label{Eq.9}
\end{equation}
where $\lambda_{rec}$, $\lambda_{cond}$, $\lambda_{w}$, and $\lambda_{emo}$ are loss weights. The gradients can be backpropagated via the reparameterization trick~\cite{kingma2013auto}.

We refer to $\mathcal{L}_{rec}$ as the reconstruction term, which can be also written as MSE loss between ground-truth and reconstructed expression coefficients: $\mathcal{L}_{rec}=\left\|{\beta_{1:t}-{\hat{\beta}_{1:t}}}\right\|_2$.
The conditional regularizer $\mathcal{L}_{cond}$ is proposed to push the approximated posterior single Gaussian distribution near each emotion component of the Gaussian mixture prior. Since this term lacks a closed-form solution, we use the 1-step Monte Carlo samples to estimate the expectation over $q_{\phi_w}({w}|\beta, a)$ and $q_{\phi_z}({z}|{\beta, a})$:
\begin{align}
\tiny
\mathcal{L}_{cond} =& \mathbb{E}_{q_{\phi_w}({w}|\beta, a)q_{\phi_z}({z}|{\beta, a})}D_{KL}(q_{\phi_z}({z}|\beta, a)\|p_\delta({z}|{w},{e}))]\nonumber \\
=&\frac{1}{N}\frac{1}{M}\sum_{n=1}^N\sum_{m=1}^M\sum_{k=1}^K
\tilde{\pi}_k \log \frac{q_{\phi_z}\left(z_m \mid \beta, a\right)}{p_\delta\left(z_m \mid w_n, e=k\right)} \nonumber \\
=& \log q_{\phi_z}\left(z \mid \beta, a \right)-\sum_{k=1}^K \tilde{\pi}_k \log p_\delta\left(z \mid w, e=k\right)
\end{align}
where $\tilde{\pi}_k$ is the posterior distribution of emotion derived from $z$ and $w$:
\begin{equation}
    \tilde{\pi}_k=p_{\delta}\left(e=k \mid {z}, w\right)=\frac{{\pi_j}\mathcal{N}({z};{\mu}_{\delta}^{j}(w),\Sigma_{\delta}^{j}({w}))}{\sum_{k=1}^K{\pi_k}\mathcal{N}({z};{\mu}_{\delta}^{k}(w),\Sigma_{\delta}^{k}({w}))}
\end{equation}

$\mathcal{L}_{w}$ is the regularizer of normal distribution $w$ which reduces KL divergence between the normal posterior and the normal prior distribution as same as vanilla VAE, formulated as:
\begin{align}
   \mathcal{L}_{w} &= D_{KL}(q_{\phi_w}({w}|{\beta, a})||p({w})) \nonumber \\
    &=-\frac{1}{2}\left(
    \log\sigma_{\phi_w}^2-
    \mu_{\phi_w}^2-\sigma_{\phi_w}^2+1\right) 
\end{align}

$\mathcal{L}_{emo}$ is the regularizer of emotion which reduces the KL divergence between the $e$-posterior and the uniform prior by pushing the same emotion samples generated from the same component of the Gaussian mixture, which can be written as:
\begin{align}
\tiny
\mathcal{L}_{emo} &=\mathbb{E}_{q_{\phi_z}({z}|{\beta}, a)q_{\phi_w}({w}|{\beta}, a)}\left[D_{KL}(p_\delta({e}|{z},{w})||p({e}))\right] \nonumber \\
&=\frac{1}{M} \sum_{i=1}^{M}
D_{KL}\left(p_\delta\left(e|{z}_{i},{w}_{i}\right) || p(e)\right) \nonumber \\
&=\sum_{k=1}^{K}\tilde{\pi}_k\left(\log \tilde{\pi}_k + \log{K}\right),
\end{align}
where we also use 1-step Monte Carlo samples to estimate the expectation over $q_{\phi_z}({z}|{\beta}, a)$ and $q_{\phi_w}({w}|{\beta}, a)$.

\subsubsection{Inference and Emotion manipulation} 
In the inference stage, we predict emotional expression coefficients according to audio feature $a_{1:t}$, target emotion label $e_{tar}$, and personalized code $s_n$ in an autoregressive manner.
As shown in Fig.~\ref{fig:pipeline}, we first sample $w$ from prior normal distribution $\mathcal{N}(0, I)$. Then, our MoG mapper generates a latent code $z_{tar}$ corresponding to the target emotion, followed by the expression decoder to predict the final expression coefficients.

Benefiting from the continuous and disentangled Gaussian mixture latent space, our method can achieve emotion manipulation by mixing the different Gaussian latent codes corresponding to each emotion.
Specifically, as shown in Fig.~\ref{fig:teaser}, given the ``happy" emotion $e_{1}$ and the ``angry" emotion $e_{2}$, we generate each latent code $z_{1}$ and $z_{2}$ through the MoG mapper determined by sampled $w$ and its emotion label. Then, we can easily blend these two latent codes by changing the interpolation weight $\alpha$:
\begin{equation}
\vspace{-0.2cm}
\small
z_{12}^{\alpha}={\alpha}\mathcal{N}({z};{\mu}_{\delta}^{1}(w),\Sigma_{\delta}^{1}({w}))+(1-\alpha)\mathcal{N}({z};{\mu}_{\delta}^{2}(w),\Sigma_{\delta}^{2}({w})) 
\end{equation}

\input{sec/3_2_FVAE}

\input{sec/3_3_StyleUNet}


%% file: sec/3_2_FVAE.tex
\vspace{-0.6cm}
\subsection{Normalizing Flow based Motion Generator} \label{FVAE}

Predicting head poses, eye blinks, and gaze from input audio is particularly challenging due to the one-to-many mapping between audio and motion.
Besides, existing methods encounter the ``mean motion" problem which means the synthesized motion tends to be over-smoothing, lacking diversity, and appearing blurred in the case of wide-range head movements.
Previous work~\cite{zhang2023sadtalker} utilizes simple distribution as the prior of VAE to predict head motion from audio, often causing the encoder to generate mean motion latent code. 
In this paper, we employ the normalizing flow technique to enhance the complexity of the prior distribution, compelling the encoder to generate diverse motion latent codes, inspired by~\cite{rezende2015variational,ren2021portaspeech,ye2023geneface}. In this way, our normalizing flow-based motion generator (NFMG) effectively mitigates the "mean motion" problem.

Given the input audio $a_{1:t}$, we utilize a normalizing flow that map a normal distribution $z_{n}$ into a more complicated distribution $z_n^{\prime}$. Subsequently, the latent representation is decoded into motion-related coefficients $\{\rho\}_{t=1}^T\in\mathbb{R}^{12}$. This process can be written as follows
\begin{align}
\vspace{-0.5cm}
    p_\psi(z_n^{\prime})=p(z_n)\Big|det\frac{\delta F_\psi}{\delta z_n}\Big|, \quad p(z_n)\sim\mathcal{N}(0,{I})\\
    p_{\psi,\theta_{\rho}}({\rho},{z_n},{a})=p_\psi(z_n^{\prime})p_{\theta_{\rho}}({\rho}|{z_n^{\prime}},{a}), 
\vspace{-0.6cm}
\end{align}
where $\psi$ and $\theta_{\rho}$ are the model parameters of normalizing flow $F_\psi$ and decoder $\mathcal{D}_{\theta_{\rho}}$. 

\noindent\textbf{Architecture.}
As illustrated in Fig.~\ref{fig:training_pipe}. (b), the proposed NFMG comprises a motion encoder, a Transformer normalizing flow, and a motion decoder.
The motion encoder $\mathcal{E}_{\epsilon}$ and the motion decoder $\mathcal{D}_{\theta_{\rho}}$ closely resemble the GMEG encoder and decoder, with the encoder featuring only one linear layer to approximate the posterior distribution $q_\epsilon(z_n|\rho, a)$.
To enhance the prior distribution of motion, we introduce a Transformer flow $F_\psi$, which is constructed by a series of $N$ invertible Transformer-based nonlinear mappings $F_\psi=F_{\psi_1}(F_{\psi_2}(...F_{\psi_N}))$ parameterized by $\psi=\{\psi_n\}_{n=1}^N$. 
Specifically, each component mapping $F_{\psi_n}$ contains two substeps: an affine coupling layer ($A$) and a flip operation. Given the latent $z_n=[z_{n1}, z_{n2}]$, the affine coupling layer aims to affinely transform half of the input elements $z_{n1}$ based on the values of the other half $z_{n2}$~\cite{henter2020moglow}. 
To employ a more powerful nonlinear transformation, we utilize 4 Transformer encoder layers as our affine coupling layer to calculate the scale $s_z$ and shift $t_z$ of $z_{n1}$, which is different from existing works~\cite{kingma2018glow,lee2020nanoflow,ren2021portaspeech}.
The flip operation ensures that after a sufficient number of flow steps, all variables can be nonlinearly transformed by reversing the ordering of the features.

\noindent\textbf{Training.} 
We utilize ELBO loss to train our NFMG. Additionally, we introduce a velocity loss to constraint temporal consistency. The loss function can be formulated as:
\begin{align}
\vspace{-0.5cm}
\mathcal{L}_{\mathrm{m}}(\epsilon, \psi, \phi_{\rho})=&\mathbb{E}_{q_\epsilon(z_n^{\prime}|\rho,a)}[\log p_{\theta_{\rho}}({\rho}|z_n^{\prime},a)]\nonumber\\
-&D_{KL}(q_\epsilon(z_n^{\prime}|\rho,a)||p_\psi(z_n^{\prime}))\nonumber\\
=&\left\|{\rho_{1:t}-{\hat{\rho}_{1:t}}}\right\|_2\nonumber\\
-&\mathbb{E}_{q_\epsilon(z_n^{\prime}|\rho,a)}[\log q_\epsilon(z_n^{\prime}|{\rho},a)-\log p_\psi(z_n^{\prime})]
\label{Eq.3}
\vspace{-0.8cm}
\end{align}
where the expectation of $q_\epsilon(z_n^{\prime}|{\rho_t},a)$ can be estimated by Monte-Carlo method.

However, merely incorporating normalizing flow proves inadequate for generating diverse motions across different data distributions. This is due to the motion acquired from a single video is usually insufficient for synthesizing diverse movements. Therefore, we pre-train our NFMG on VoxCeleb2~\cite{chung2018voxceleb2}, a large dataset with diverse head and eye movements, to enhance the generalization of motion. 
So that we can make full use of the potential of the normalizing flow to fit complex data distributions and alleviate the “mean head motion” problem.

\noindent\textbf{Inference.} 
As shown in Fig.~\ref{fig:training_pipe}. (b), our Transformer flows first map a latent variable $z_n$ sampled from the prior normal distribution into $z_n^{\prime}$. Then we pass this latent variable with audio feature $a_{1:t}$ into the motion decoder to generate diverse and realistic motion coefficients $\hat{\rho}_{1:t}$ autoregressively similar to Section~\ref{GMVAE}.
Since the $z_n$ is randomly sampled, we can generate different head poses, eye blinks, and gazes given the same speech. Furthermore, we can also generate diverse and realistic motions given the different audio due to the powerful generative ability of the proposed NFMG.

%% file: sec/3_3_StyleUNet.tex
\vspace{-0.2cm}
\subsection{Emotion-guided Head Generator} \label{StyleUNet}

\begin{figure*}
  \centering
  \includegraphics[width=0.93\textwidth]{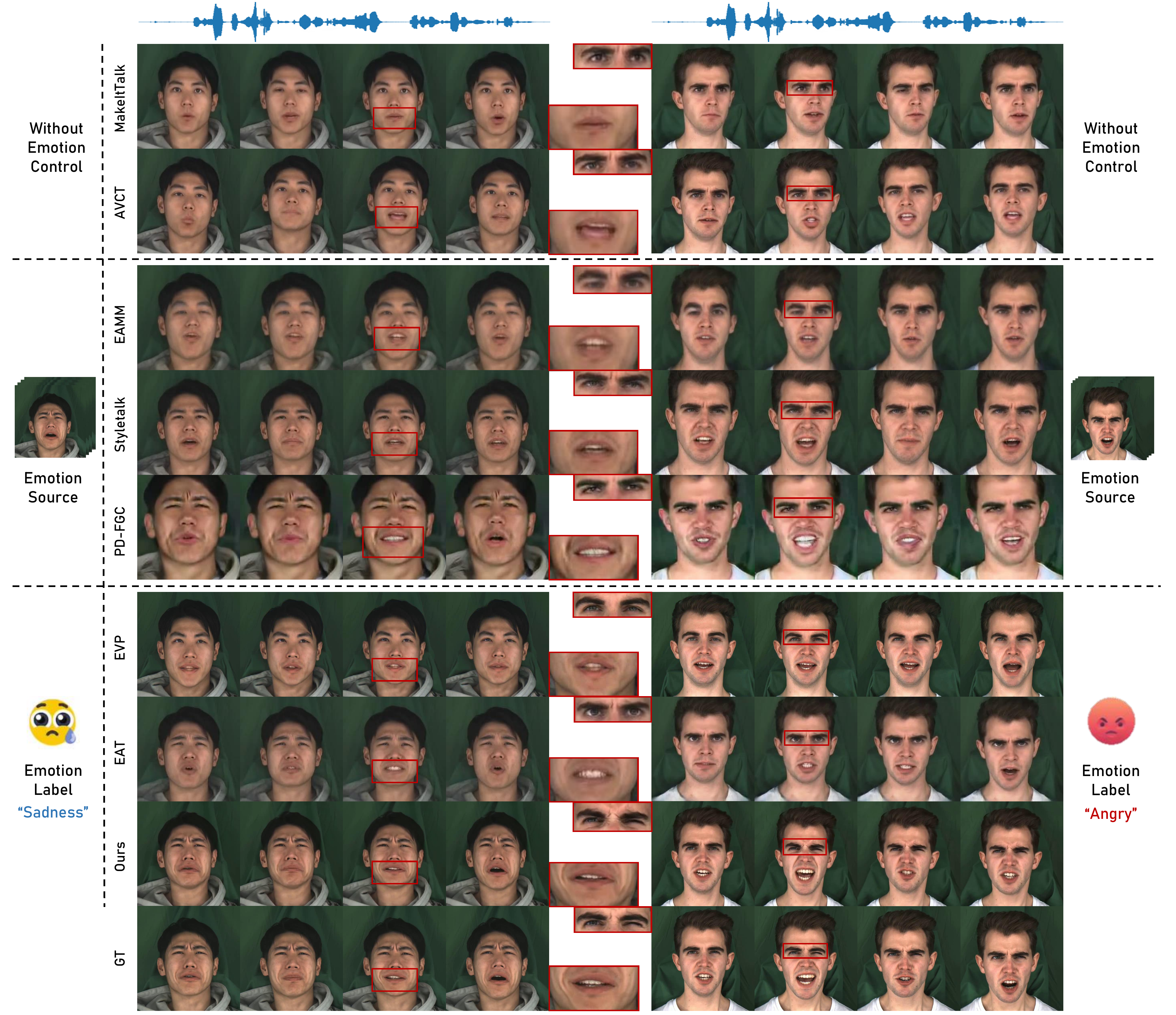}
  \vspace{-0.4cm}
  \caption{Qualitative comparison for emotional talking video portraits on the two cases in the MEAD test dataset. 
  The emotion categories of the videos are happy (left) and angry (right). The bottom row shows ground-truth frames. Since EAMM~\cite{ji2022eamm} and EAT are one-shot methods, we choose the same reference image used in EAT~\cite{gan2023efficient} to generate target videos for them. }
  \label{fig:visualcomparsion}
\vspace{-0.5cm}
\end{figure*}

After generating emotional expression and other motion coefficient sequences from input audio (Section~\ref{GMVAE} and Section~\ref{FVAE}), we render them to images using the fixed shape and texture coefficients of the target person (Section~\ref{Preliminaries}). Next, we will generate photo-realistic images from synthetic 3DMM renderings through an image-to-image translation generator. 
We utilize the style-based generator, StyleUNet~\cite{wang2023styleavatar}, which can reconstruct the personalized style of a target identity. However, owing to the highly coupled latent space of StyleUNet, it struggles to explicitly control the emotion that we want.  Therefore, we introduce an Emotion Mapping Network (EMN) to branch each emotion type corresponding to the available sub-domain, motivated by~\cite{choi2020stargan}. 

\noindent\textbf{Emotion Mapping Network.} Given a latent code $z_{style}$ and an emotion label, our EMN generates a style code $\mathbf{s}_{style}=\mathcal{M}_{k}({z_{style}})$, where $\mathcal{M}_{k}(.)$ denotes the output corresponding to the emotion type $k$, and $\mathcal{M}$ consist of an MLP with two shared layers and $K$ multiple unshared layers (similar to the emotion numbers).
Consequently, we can utilize $z_{style}$ to control emotion-related details, such as facial wrinkles. 
Please refer to Appendix~\ref{appendix:A1_EMN_detail} of our supplementary materials for more details.

%% file: sec/4_experiment.tex
\vspace{-0.2cm}
\section{Experiments}
\label{sec:experiments}

\input{table/comparsion}
\input{table/compare_cremad}

In this section, we first describe the experimental setup of our approach in Section~\ref{sec:4_1_setup}: dataset and implementation details, baseline method, and evaluation metrics. Subsequently, we present 
the comparison results in Section~\ref{sec:4_2_comparsion}.
Finally, we show results of the ablation study in Section~\ref{sec:4_4_ablation}.
\vspace{-0.5cm}
\subsection{Experimental Setup}\label{sec:4_1_setup}
\noindent\textbf{Dataset and Implementation Details.} 
We conduct emotional experiments on the commonly used talking head datasets, MEAD~\cite{wang2020mead} and CREMA-D~\cite{cao2014crema}. 
As for non-emotional target-person experiments, we utilize the video samples provided by LSP~\cite{lu2021live} with 80\%/20\% for training and testing. 
Notably, we set the component of our Gaussian mixture $K$ to 1 for non-emotional experiments, representing a normal distribution. Similarly, the number of unshared layers in EMN is set to 1. 
To learn diverse and large motion changes, we select about 20k videos from the VoxCeleb2~\cite{chung2018voxceleb2} to pre-train our NFMG. Then, we fine-tune this diverse motion prior to a specific person taking a few minutes.
More details of dataset and implementations are provided in Appendix~\ref{appendix:A2_data_train_details} of the supplementary materials.

\noindent\textbf{Baseline.} We compare our method with: (1) emotion-agnostic talking video generation methods: MakeItTalk~\cite{zhou2020makelttalk}, Wav2Lip~\cite{prajwal2020lip}, and AVCT~\cite{wang2022one}. (2) emotion-controllable talking video generation methods: EAMM~\cite{ji2022eamm}, Styletalk~\cite{ma2023styletalk}, PD-FGC~\cite{wang2023progressive}, and EAT~\cite{gan2023efficient}.
Besides, we additionally compare our results on the CREMA-D dataset with Vougioukas et.al~\cite{vougioukas2020realistic}, Diffused Heads~\cite{stypulkowski2024diffused} and ETK~\cite{eskimez2021speech}.
For motion-controllable talking video generation, we compare our method with several state-of-the-art motion-controllable methods: FACIAL~\cite{zhang2021facial}, LSP~\cite{lu2021live}, Audio2Head~\cite{wang2021audio2head} and SadTalker~\cite{zhang2023sadtalker}, focusing on high-quality portrait generation with natural and diverse motion.

\noindent\textbf{Evaluation Metrics.}
We use the following metrics used in EAT~\cite{gan2023efficient} to measure the visual quality and audio-visual synchronization for all quantitative experiments. For emotional talking video comparison, we employ $Acc_{emo}$ to assess the emotional accuracy of synthesized videos. 
To measure the diversity of generated head motion, we adopt Beat Alignment (\textbf{BA}) and Diversity (\textbf{Div}) metrics used in SadTalker~\cite{zhang2023sadtalker}, as well as Percent of Correct Motion (\textbf{PCM}) mentioned in BEAT~\cite{liu2022beat}.
Detailed descriptions of metrics are available in Appendix~\ref{appendix:B_metrics} of the supplementary materials.

\vspace{-0.2cm}
\subsection{Comparison Results} \label{sec:4_2_comparsion}

\begin{figure}[t]
  \centering
  \includegraphics[width=0.48\textwidth]{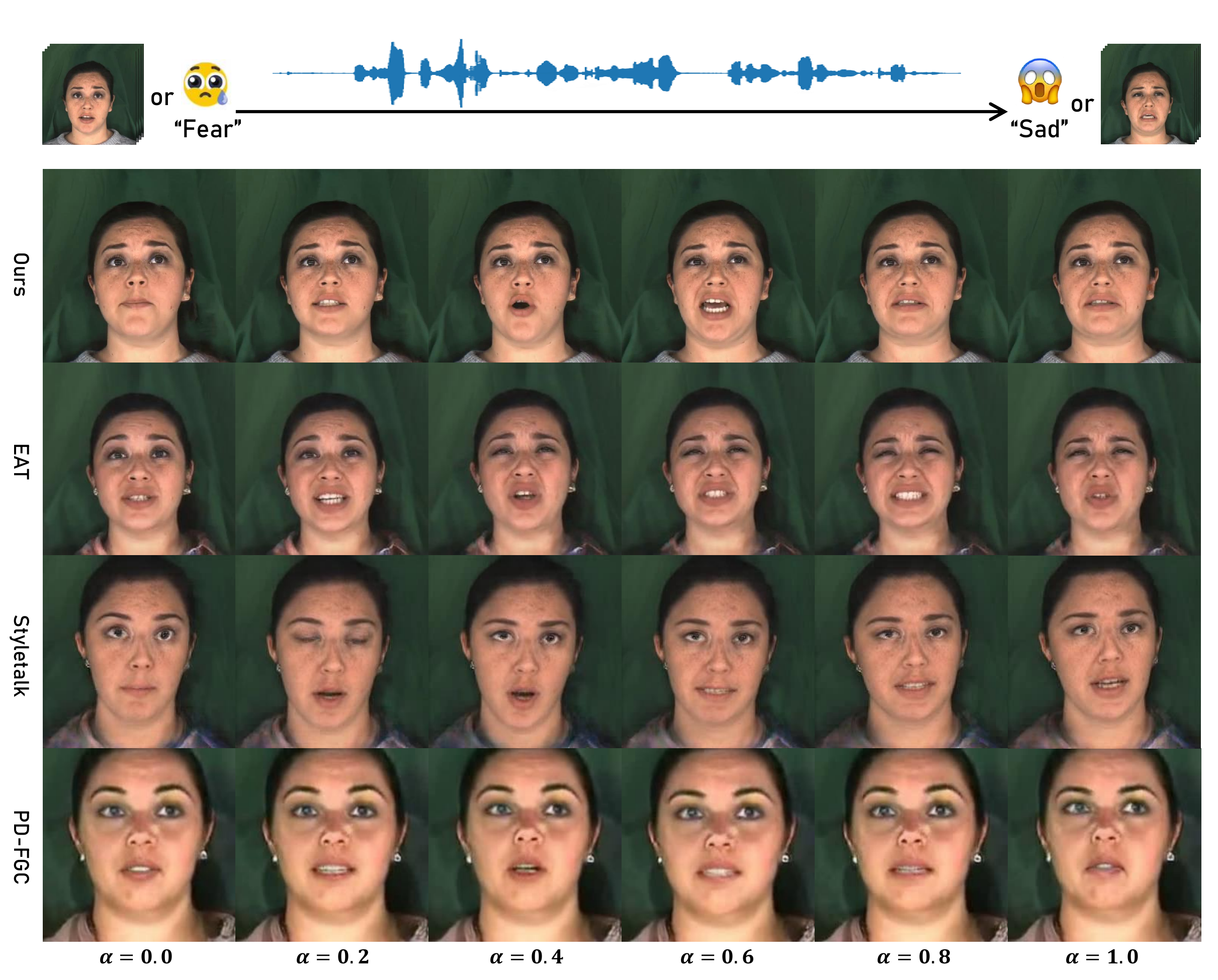}
  \vspace{-0.3cm}
  \caption{Qualitative results of the emotion interpolation comparison. 
  For PD-FGC~\cite{wang2023progressive} and Styletalk~\cite{ma2023styletalk}, we manipulate expressions between the source emotion video and the target emotion video. For EAT~\cite{gan2023efficient}, we achieve this by transitioning between the source emotion label and the target emotion label.}
  \label{fig:interp_compare}
\vspace{-0.6cm}
\end{figure}

\noindent\textbf{Comparison of Emotion Generation.} 
For quantitative comparison, we adopt the experimental setup and evaluation method utilized by EAT~\cite{gan2023efficient} on the public-available MEAD and CREMA-D test set.  
As shown in Table~\ref{tab:tablecompare} and Table~\ref{tab:tablecompare_cremad}, GMTalker achieves the best overall visual quality and emotion accuracy, demonstrating the superior quality of disentangled emotion latent space learned by GMEG. In terms of the \textbf{Sync} score, our method shows comparable performance with other methods. Please note that a higher sync score does not invariably guarantee better results, as this metric can be overly sensitive to audio and the SyncNet model is trained only on neutral videos. 
The higher scores achieved by Wav2Lip~\cite{prajwal2020lip} and EAT~\cite{gan2023efficient} may be attributed to their overfitting of the pretrained SyncNet model or utilizing synchronization loss.


Moreover, we perform qualitative comparisons with serval state-of-the-art methods using the test sets of the MEAD and CREMA-D (see Appendix \ref{appendix:comparision} in the supplementary materials). Illustrated in Fig.~\ref{fig:visualcomparsion}, our method excels in generating high-fidelity and faithful emotional talking video portraits. 
While these methods either fail to express desired emotion or exhibit artifacts in the mouth region, GMTalker remains comparatively more faithful to the ground truth expressions, including maintaining natural mouth shapes aligned with the input speech. Additionally, it produces detailed expressions with personalized speaking styles, such as realistic wrinkles in the face and around the eyes.

\noindent\textbf{Comparison of Emotion Interpolation.} 
To illustrate the continuity and decoupling characteristics of our Gaussian mixture latent space learned by GMEG, we conduct an emotion interpolation study comparing our GMTalker with several state-of-the-art methods: EAT~\cite{gan2023efficient}, Styletalk~\cite{ma2023styletalk} and PD-FGC~\cite{wang2023progressive}. As shown in Fig.~\ref{fig:interp_compare}, given the same driving audio, we obtain the image sequences of different methods by interpolating between the corresponding emotion features extracted from the source emotion ``Fear" and the target emotion ``Sad''. 
For Styletalk and PD-FGC, which extract emotion features from additional emotion videos, the generated facial emotion dynamics lack smooth emotion transition and fail to achieve the desired emotion states. 
EAT generates emotional embeddings through its deep emotional prompts, resulting in relatively continuous facial expression dynamics. However, it may present ambiguous emotional states: the generated target expression ``Sad" more closely resembles confusion or fear.
In contrast, by interpolating in our continuous and disentangled Gaussian mixture latent space, we can smoothly transition from the source expression to the target expression while preserving the accuracy of the emotion.

\input{table/compare_interp}

To quantitatively validate the quality of intermediate facial expression and the smoothness of the emotional transition video, we introduce two new metrics inspired by~\cite{karras2020analyzing, zhang2023diffmorpher}.
Emotion Perceptual Path Length (\textbf{E-PPL}, $\downarrow$) serves as an indicator of the emotional smoothness and consistency of the generated transition video, and Emotion Perceptual Distance Variance (\textbf{E-PDV}, $\downarrow$) serves as a natural measure of the homogeneity of the emotion video transition rate. The details of these two metrics are shown in Appendix~\ref{appendix:B_interp_metrics} of the supplementary materials.

Besides, we adopt the SyncNet score to evaluate the audio-visual synchronization of the generated emotional transition video. The quantitative results of all approaches are presented in Table~\ref{tab:tablecompareinterp}. Our methods achieve significantly lower \textbf{E-PPL} and \textbf{E-PDV} than others, showing smoother emotion transitions. 
On the other hand, our GMTalker has a higher Sync score, demonstrating more accurate lip synchronization when the emotions are changed. 

\begin{figure}[t]
  \centering
  \includegraphics[width=0.48\textwidth]{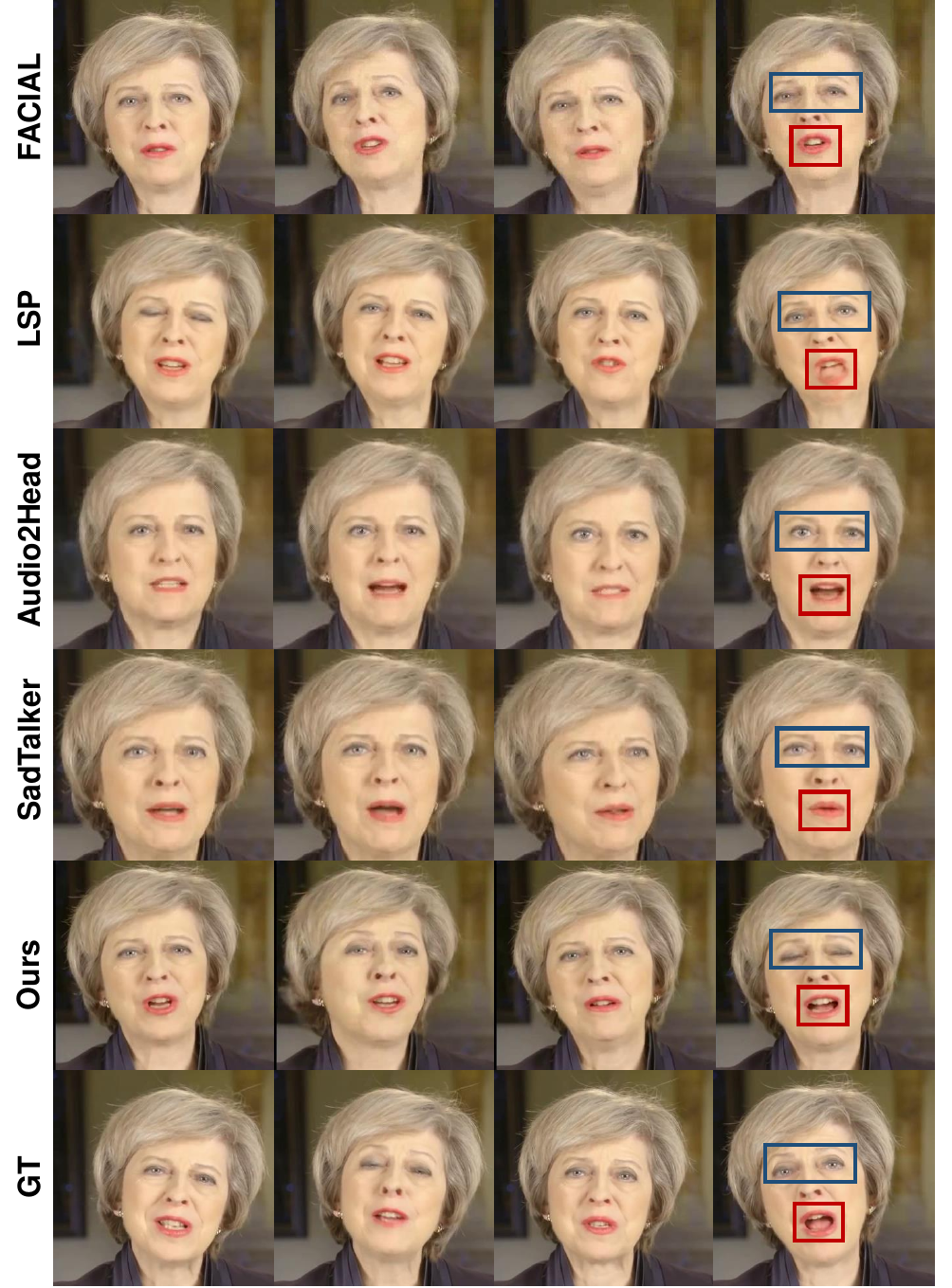}
  \vspace{-0.3cm}
  \caption{Qualitative comparison for motion-controllable methods on the one test sample from LSP~\cite{lu2021live}.}
  \label{fig:singlecomparsion}
\vspace{-0.5cm}
\end{figure}

\input{table/compareLSP_new_metric}

\noindent\textbf{Comparison of Motion Generation.} 
We compare our approach with several state-of-arts pose-controllable methods: FACIAL~\cite{zhang2021facial}, LSP~\cite{lu2021live}, Audio2Head~\cite{wang2021audio2head} and SadTalker~\cite{zhang2023sadtalker}, on test video samples from LSP. As depicted in Table~\ref{tab:tablecomparesingle_new}, our method shows outstanding performance in terms of \textbf{Div}, \textbf{BA}, and \textbf{PCM}. 
Meanwhile, we also achieve the best performance on overall visual quality and lip sync metrics. 
This suggests that the latent space learned by normalizing flow can represent complex motion distributions derived from pretrained models. 
Qualitative comparisons are shown in Fig.~\ref{fig:singlecomparsion}. 
Our approach excels in generating diverse head motions, natural eye blinks, and accurate mouth shapes with audio, and visual quality compared with other methods.
For more detailed comparison results please refer to Appendix~\ref{appendix:comparision} in our supplement materials.

\input{table/ablation_emn}
\input{table/ablation_flow}

\begin{figure}[t]
  \centering
  \includegraphics[width=0.48\textwidth]{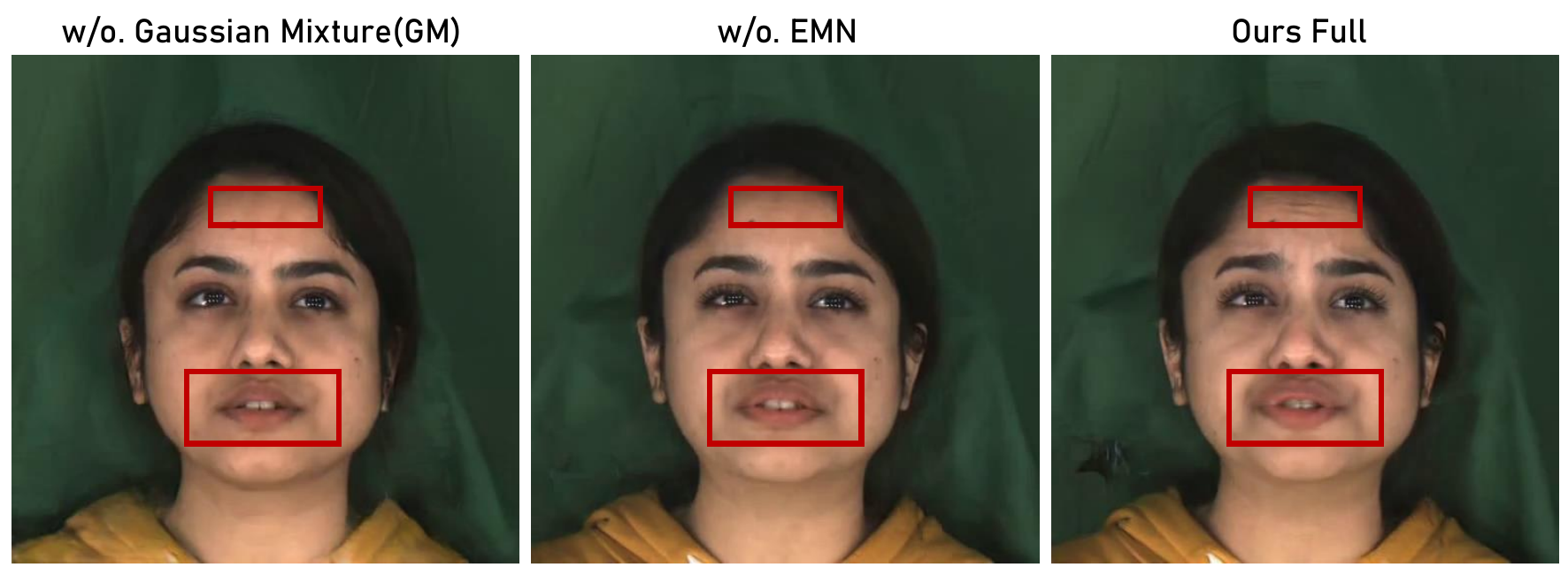}
  \vspace{-0.3cm}
  \caption{Qualitative results of ablation study for Gaussian mixture (GM) prior and EMN. From left to right, the results are w/o GM, with GM but w/o EMN, and our full model.}
  \label{fig:AblationEMN}
\vspace{-0.3cm}
\end{figure}

\vspace{-0.1cm}
\subsection{Ablation Study} \label{sec:4_4_ablation}
To demonstrate the effectiveness of our designed choice, we perform ablation studies with the following three alternative modules: 1) w/o Gaussian mixture prior in GMEG (w/o GM), where we replace the Gaussian mixture distribution prior with an unimodal Gaussian prior, 2) w/o EMN, where we use 4-layers MLP as mapping network without emotion-guided,
3) w/o normalization flow, where we use the normal distribution as the prior of the motion generator.

\paragraph{The Ablation of GMEG and EMN}
We conduct the ablation of GMEG and EMN on the MEAD test set. We compare the visual quality, audio-visual synchronization, and emotional accuracy metrics before and after removing the Gaussian mixture prior in GMEG and EMN, respectively.
As shown in Table~\ref{tab:tableemn}, all components can improve video quality and emotion accuracy. 
In particular, when removing the constraints on the Gaussian mixture distribution, there is a significant decline in the emotional accuracy of the generated videos, demonstrating the effectiveness of our disentangled emotion latent space.
A qualitative case in generating ``Fear'' expression is illustrated in Fig.~\ref{fig:AblationEMN}.
It can be observed that when replacing the Gaussian mixture distribution with normalization distribution or removing the EMN module, the model can hardly generate the desired expression and facial details, including wrinkles, mouth shapes, and eye gaze.

\paragraph{The Ablation of NFMG}
For the ablation of NFMG, we focus on the \textbf{Div}, \textbf{BA}, and \textbf{PCM} metrics. 
As shown in Table~\ref{tab:tableablationflow}, the quantitative results demonstrate that our normalizing flow with pre-training can provide richness motion prior and generate diverse and wide-range motions. 
For more details please refer to Appendix~\ref{appendix:E_ablation} of our supplement materials.

%% file: table/comparsion.tex
\begin{table*}[t]
\centering
\caption{Quantitative comparisons with the state-of-the-art methods on MEAD~\cite{wang2020mead}. Please note that the sync value for emotional talking video portraits may be inaccurate as the SyncNet model is trained only with neutral videos.}
\vspace{-0.3cm}
\small
\setlength\tabcolsep{3.5pt}
\begin{tabular}{c|c|c c c |c c c|c|ccc}
\toprule

\multicolumn{1}{c}{\multirow{2}{*}{Method}} & 
\multicolumn{1}{c}{\multirow{2}{*}{Emo Source}} &\multicolumn{3}{c}{Visual Quality} & \multicolumn{3}{c}{Lip Synchronization} & \multicolumn{1}{c}{Emo Accuracy}  & \multicolumn{3}{c}{Output} \\

\cmidrule{3-12} 
\multicolumn{1}{c}{} & \multicolumn{1}{c}{} & 
 PSNR$\uparrow$ & SSIM$\uparrow$ & \multicolumn{1}{c}{}
{FID$\downarrow$} & Sync$\uparrow$ & M-LMD$\downarrow$ &\multicolumn{1}{c}{} {F-LMD$\downarrow$} & $\operatorname{Acc}_{emo}\uparrow$  &
Pose &  Eye & Emo \\ 

\midrule
MakeItTalk~\cite{zhou2020makelttalk} & N./A. & 18.79 & 0.55 & 51.88 & 5.28 & 3.61 & 4.00 & 15.23  & generate & \usym{2717} & \usym{2717}\\
Wav2Lip~\cite{prajwal2020lip} & N./A. & 19.12 & 0.57 & 67.49 & \textbf{8.97} & 3.11 & 3.71 & 17.87  & \usym{2717} & \usym{2717} & \usym{2717}\\
AVCT~\cite{wang2022one} & N./A. & 18.43 & 0.54 & 39.18 & 6.02 & 3.82 & 4.33 & 15.64 & generate & \usym{2717} & \usym{2717} \\
\midrule
EAMM~\cite{ji2022eamm} & Video &  20.55 & 0.66 & 22.38 & 6.62 & 2.19 & 2.55 & 49.85  & transfer & \usym{2717} & \usym{2713}\\
Styletalk~\cite{ma2023styletalk} & Video &  19.27 & 0.64 & 49.73 & 5.29 & 2.67 & 3.45 & 45.41 & transfer & \usym{2717} & \usym{2713}\\
PD-FGC~\cite{wang2023progressive} & Video &  20.97 & 0.65 & 39.52 & 6.29 & 1.80 & 2.15 & 36.09  & transfer & transfer & \usym{2713}\\
\midrule
EAT~\cite{gan2023efficient} & Label &  21.75 & 0.68 & 19.69 & 8.28 & 2.25 & 2.47 & 75.43  & transfer & \usym{2717} & \usym{2713}\\
GMTalker (Ours) & Label &  \textbf{24.86} & \textbf{0.80}  & \textbf{13.49} & 7.34 & \textbf{1.23}& \textbf{1.48}  & \textbf{83.92} 
& generate & generate & \usym{2713} \\
Ground Truth & N./A. & $\infty$ & 1.00 & 0 & 7.76 & 0.00 & 0.00 & 84.37 & - & - & - \\
\bottomrule
\end{tabular}
\vspace{-0.2cm}
\label{tab:tablecompare}
\end{table*}

%% file: table/compare_cremad.tex
\begin{table*}[t]
\centering

\caption{Quantitative comparisons with the state-of-the-art methods on CREMA-D~\cite{cao2014crema}.}
\vspace{-0.3cm}
\small
\setlength\tabcolsep{3.0pt}
\begin{tabular}{c|c|c c c |c c c|c|ccc}
\toprule
\multicolumn{1}{c}{\multirow{2}{*}{Method}} & 
\multicolumn{1}{c}{\multirow{2}{*}{Emo Source}} &\multicolumn{3}{c}{Visual Quality} & \multicolumn{3}{c}{Lip Synchronization} & \multicolumn{1}{c}{Emo Accuracy}  & \multicolumn{3}{c}{Output} \\

\cmidrule{3-12} 
\multicolumn{1}{c}{} & \multicolumn{1}{c}{} & 
 PSNR$\uparrow$ & SSIM$\uparrow$ & \multicolumn{1}{c}{}
{FID$\downarrow$} & Sync$\uparrow$ & M-LMD$\downarrow$ &\multicolumn{1}{c}{} {F-LMD$\downarrow$} & $\operatorname{Acc}_{emo}\uparrow$  &
Pose &  Eye & Emo \\ 
\midrule
MakeItTalk~\cite{zhou2020makelttalk} & N./A. & 21.98 &  0.67 & 29.99 & 3.42 & 3.08 & 3.27 &  17.21 & generate & \usym{2717} & \usym{2717}\\ 
Vougioukas et.al~\cite{vougioukas2020realistic} & N./A. & 22.11 &  0.68 & 34.93 & 5.01 &2.10 & 2.63 &  26.81 & \usym{2717} & \usym{2717} & \usym{2717}\\ 
AVCT~\cite{wang2022one} & N./A. & 20.65 &  0.63 & 23.69 & 5.42 & 2.87 & 3.77 & 14.87 & generate & \usym{2717} & \usym{2717} \\ %
Diffused Heads~\cite{stypulkowski2024diffused} & N./A. & 22.16 & 0.68 & 20.49 & 5.13 & 2.31 & 2.91 & 28.54 & \usym{2717} & \usym{2717} & \usym{2717} \\ 
\midrule
EAMM~\cite{ji2022eamm} &Video & 21.21 &  0.66 & 39.00 & 3.75 & 2.64 & 3.16 & 21.12  & transfer & \usym{2717} & \usym{2713}\\ 
Styletalk~\cite{ma2023styletalk}&Video & 23.78 &  0.75 & 13.98 & 3.55 & 2.08 & 2.10 & 52.32 & transfer & \usym{2717} & \usym{2713}\\
PD-FGC~\cite{wang2023progressive} &Video & 23.82 &  0.73 & 24.86 & 4.77 & 1.55 & 1.85 & 46.22 & transfer & transfer & \usym{2713} \\ 
\midrule
ETK~\cite{eskimez2021speech} & Label & 23.34 &  0.72 & 18.08 & 5.42 & 1.81 & 2.43 & 63.05 & \usym{2717} & \usym{2717} & \usym{2713} \\
EAT~\cite{gan2023efficient} &Label&  21.66 & 0.66 & 20.78 & 5.78 & 2.62 & 2.89 & 46.09 & transfer & \usym{2717} & \usym{2713} \\ 
GMTalker (Ours) &Label&  \textbf{24.16} & \textbf{0.77}  & \textbf{9.24} & \textbf{6.80} & \textbf{1.43}& \textbf{1.59}  & \textbf{82.91} & generate & \usym{2713} & \usym{2713}\\ %
Ground Truth & N./A. & $\infty$ & 1.00 & 0 & 7.76 & 0.00 & 0.00 & 98.42 &-&-&- \\ 
\bottomrule
\end{tabular}

\vspace{-0.5cm}
\label{tab:tablecompare_cremad}
\end{table*}

%% file: table/compare_interp.tex
\begin{table}[t]
\centering

\caption{Quantitative comparison of the emotion interpolation study.}
\vspace{-0.3cm}
\begin{tabular}{lcccc}
\toprule

Method & Emo Source & E-PPL$\downarrow$ & E-PDV$\downarrow$  & Sync$\uparrow$  \\ 
\midrule
PD-FGC~\cite{wang2023progressive} & Video & 21.08 & 44.95  & 5.30 \\
Styletalk~\cite{ma2023styletalk} & Video & 12.84 & 15.35  & 5.61\\
EAT~\cite{gan2023efficient} & Label & 8.89 & 7.28 & 6.64 \\
Ours & Label &  \textbf{6.64} & \textbf{6.86}   & \textbf{6.74} \\
\bottomrule

\end{tabular}
\vspace{-0.6cm}
\label{tab:tablecompareinterp}
\end{table}

%% file: table/compareLSP_new_metric.tex
\begin{table*}[t]
\centering
\caption{Quantitative comparisons with state-of-the-art pose-controllable method on LSP~\cite{lu2021live} test samples. We evaluate SadTalker~\cite{zhang2023sadtalker} in the one-shot settings, and others in person-specific settings.}
\vspace{-0.3cm}
\begin{tabular}{l|c c c |c c c|c c c}
\toprule
\multicolumn{1}{c}{\multirow{2}{*}{Method}} & 
\multicolumn{3}{c}{Visual Quality} & \multicolumn{3}{c}{Lip Synchronization} & \multicolumn{3}{c}{Motion Diversity} \\
\cmidrule{2-10} \multicolumn{1}{c}{}  & PSNR$\uparrow$ & SSIM$\uparrow$ & \multicolumn{1}{c}{} 
{FID$\downarrow$} & Sync$\uparrow$ & M-LMD$\downarrow$ & \multicolumn{1}{c}{} {F-LMD$\downarrow$} & BA$\uparrow$ & Div$\uparrow$ & PCM$\uparrow$ \\ \midrule
FACIAL~\cite{zhang2021facial}  &  19.22 & 0.62  & 35.14 & 3.99 & 1.82 & 2.83 &0.242&2.182 & 0.411\\
LSP~\cite{lu2021live}  & 19.48 & 0.62 & 46.45 & 4.93 & 1.99 & 2.84 & 0.252 & 2.173 & 0.432\\
Audio2Head~\cite{wang2021audio2head} & 16.97 & 0.53 &56.06 & 6.94 & 2.47 & 3.89 & 0.259 & 2.182 & 0.433  \\ 
SadTalker~\cite{zhang2023sadtalker} & 17.23 & 0.54 & 50.15 & \textbf{7.82} & 2.51 & 3.84 & 0.263 & 1.934 & 0.408 \\
Ours & \textbf{19.80} & \textbf{0.64} & \textbf{30.91} & \textbf{7.82} & \textbf{1.79}& \textbf{2.75} & \textbf{0.290} & \textbf{2.201}& \textbf{0.449}\\
Ground Truth & $\infty$ & 1.00 & 0 & 8.53 & 0.00 & 0.00 & 0.271 &  2.151 & 0.435 \\
\bottomrule
\end{tabular}

\vspace{-0.3cm}
\label{tab:tablecomparesingle_new}
\end{table*}

%% file: table/ablation_emn.tex
\begin{table}[t]
\centering

\caption{Quantitative ablation study results for GMEG and EMN.}
\vspace{-0.3cm}
\setlength\tabcolsep{3.0pt} 
\begin{tabular}{lccccc}
\toprule
Ablation & PSNR/SSIM$\uparrow$ & FID$\downarrow$  & M/F-LMD$\downarrow$  & Sync$\uparrow$  & $Acc_{emo}$ $\uparrow$  \\ 
\midrule
w/o GM & 23.71/0.77 & 14.64  & 1.61/1.85 & 7.23 & 65.61 \\
w/o EMN & 24.82/0.78 & 12.95 & 1.62/1.81 & 7.30 & 72.33 \\
Ours Full &  \textbf{25.18}/\textbf{0.81} & \textbf{11.86}   & \textbf{1.52}/\textbf{1.76} & \textbf{7.41} & \textbf{77.73} \\
\bottomrule
\end{tabular}
\vspace{-0.3cm}
\label{tab:tableemn}
\end{table}

%% file: table/ablation_flow.tex
\begin{table}[t]
\centering
\caption{Quantitative results of the ablation study for normalizing flow motion generator on test samples from LSP.}
\vspace{-0.3cm}
\setlength\tabcolsep{6.0pt} 
\begin{tabular}{lccc}

\toprule Ablation & BA$\uparrow$ & Div$\uparrow$ & PCM$\uparrow$ \\ 
\midrule
w/o normalizing flow & 0.244 & 2.101 & 0.427 \\
w/o pre-train & 0.272 & 2.189 &0.438 \\
Ours Full &  \textbf{0.293} & \textbf{2.223} & \textbf{0.447}\\
\bottomrule

\end{tabular}
\vspace{-0.4cm}
\label{tab:tableablationflow}
\end{table}

%% file: sec/5_conclusion.tex
\vspace{-0.2cm}
\section{Discussion and Conclusion}
\label{sec:diss}

\paragraph{Limitation} Although our method has demonstrated its superiority compared with existing emotional talking video portrait methods, there are still several limitations: (1) our method relies on high-quality videos containing rich emotional content, the capturing of which comes with certain challenges; (2) our method still describes limited emotions subjecting to the eight categories in the dataset and need to train on the target person.

\paragraph{Potential Social Impact}
As our method is capable of producing realistic emotional talking portraits from monocular videos, there is a potential for its application in creating deceptive talking videos, which should be addressed carefully before its deployment.

\paragraph{Conclusion} 
In this paper, we present GMTalker which can generate high-fidelity and faithful emotional talking video portraits with diverse motions.
To achieve precise emotion control and continuous emotion transition, we propose the GMEG to construct a continuous and disentangled Gaussian mixture latent space. Then, NFMG is proposed to alleviate the ``mean motion" problem and predict diverse head poses, eye blinks, and gazes. Finally, we introduce an emotion-guided head generator with the proposed EMN to generate high-quality emotional talking video portraits with personalized speaking styles. 
By incorporating GMEG, NFMG, and EMN, our method offers a unique blend of advantages, combining faithful and smooth emotion interpolation, diverse head and eye motions, and high-quality video generation.
Overall, experiments have demonstrated that our method outperforms other state-of-the-art approaches, and we believe that our Gaussian mixture latent space will inspire future research on talking head generation.

%% file: sec/suppl_1.tex
\section{More Implementation Details}
 
\subsection{Details of Emotion-guided Head Generator} \label{appendix:A1_EMN_detail}
Here, we present more details of the Emotion-guided Head Generator, which are introduced in the main paper.
\subsubsection{Shoulder Mask}
To achieve more stable results, we incorporate a shoulder mask as an additional input different from StyleUNet~\cite{wang2023styleavatar}. Specifically, given the input monocular video, we perform facial parsing using BiSeNet~\cite{yu2018bisenet} to obtain shoulder masks of each frame.
These shoulder masks are concatenated with the corresponding frames of the 3DMM renderings and together serve as input to the network. During the driving phase, stable shoulder motion is achieved by providing a reference image of the shoulder mask.
\subsubsection{Training Loss.} 
The emotion-guided head generator is trained in an adversarial way with a discriminator, which keeps the same architecture as the StyleGAN2 discriminator~\cite{karras2020analyzing}. During the training process, input with a single synthetic 3DMM rendering and an emotion label, we generate an output image. This output image is then concatenated with the 3DMM rendering as a 6-channel ``fake" input for the discriminator.
The corresponding ground-truth image and 3DMM rendering are concatenated into ``real" input.
As a result, we train our emotion-guided head generator using common L1 loss $\mathcal{L}_{1}$, perceptual loss $\mathcal{L}_{per}$, and GAN loss $\mathcal{L}_{GAN}$ following~\cite{wang2023styleavatar}, which can be formulated as:
\begin{equation}
\mathcal{L}_{render}=\mathcal{L}_{1}+\mathcal{L}_{per}+\mathcal{L}_{GAN}
\vspace{-0.2cm}
\end{equation}

\subsection{Dataset and Training Details} \label{appendix:A2_data_train_details}
We conduct experiments on MEAD~\cite{wang2020mead}, CREMA-D~\cite{cao2014crema}, and LSP~\cite{lu2021live} dataset.
MEAD is a high-quality emotional talking video dataset with 8 kinds of emotions and audio-visual recordings performed by different actors. CREMA-D contains video clips of a variety of different age groups and races uttering 12 sentences expressing six categorical emotions. We train and test our model on 6 subjects for the MEAD dataset and 10 subjects for the CREMA-D dataset. 
LSP dataset consists of five video samples featuring four different celebrities, with an average video duration of 4 minutes.
All the videos are sampled in 25 FPS, and the audio sample rate is 16 kHz. The MEAD and CREMA-D videos are cropped and resized to $512\times512$ and $256\times256$, while the LSP dataset remains $512\times512$. 

We train our GMEG for 100 epochs taking about 8-10 hours using Adam optimizer where the learning rate is $1\times10^4$, beta1 is 0.9 and beta2 is 0.999. For loss weights in Eq. 7, we empirically set the loss weight $\lambda_{rec}$ as 1.0, and $\lambda_{w}$, $\lambda_{emo}$, $\lambda_{cond}$ as 0.5.
We use the same hyperparameter settings with~\cite{wang2023styleavatar} to train the emotion-guided head generator on a specific person for about 4-6 hours.
All experiments are conducted on an NVIDIA 3090 GPU and Pytorch framework.

\begin{figure}[t]
  \centering
  \includegraphics[width=0.48\textwidth]{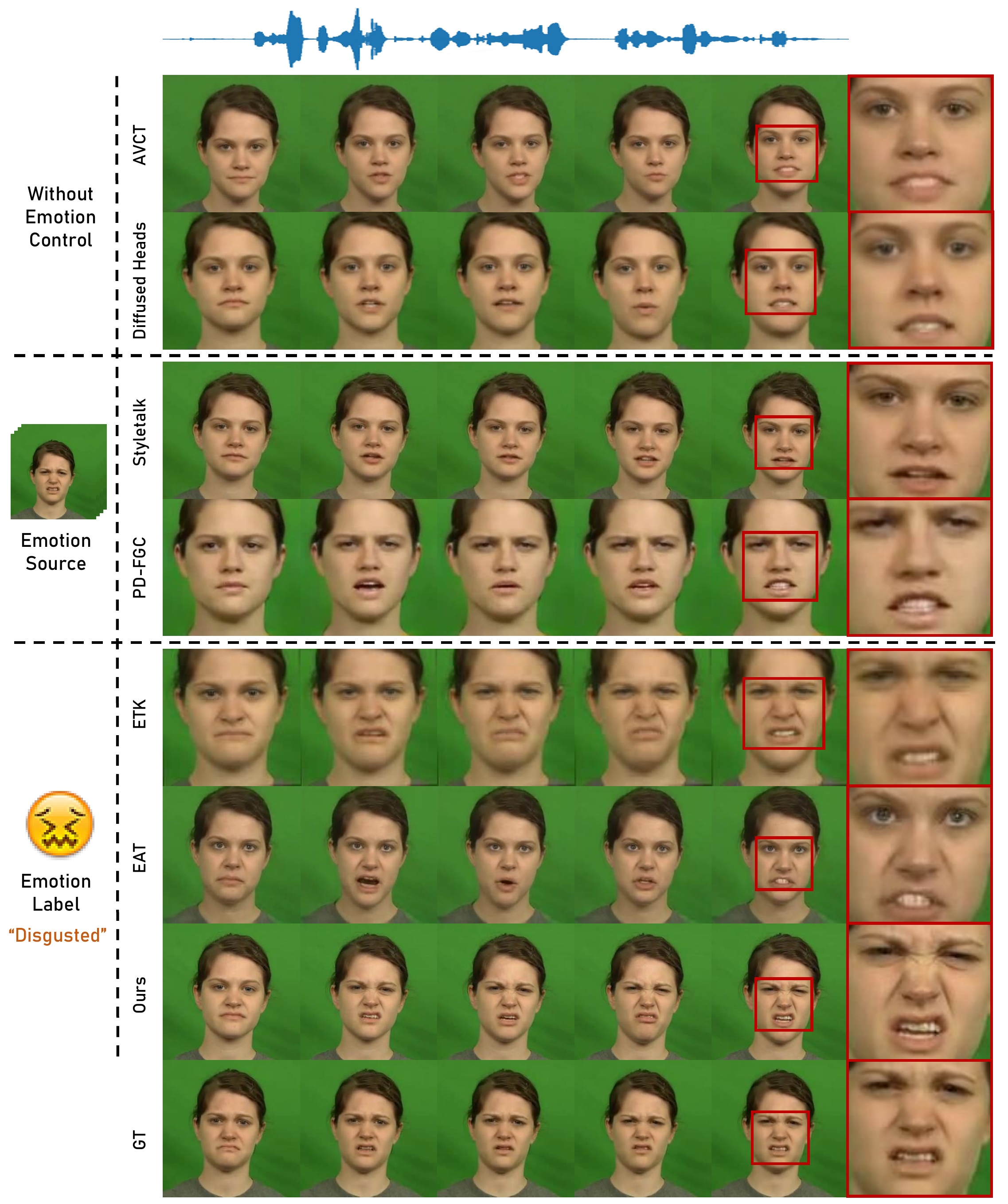}
  \vspace{-0.2cm}
  \caption{Qualitative comparison for emotional talking video portraits on the CREMA-D~\cite{cao2014crema} test dataset. 
  The emotion category of the videos is anger. The bottom row shows ground-truth frames.}
  \label{fig:cremadcomparsion}
\vspace{-0.2cm}
\end{figure}

\section{Evaluation Metric} \label{appendix:B_metrics}
Here, we provide details of the evaluation metrics used in our main paper.
\textit{Audio-visual synchronization.} We use the lip sync confidence score (\textbf{Sync}) of SyncNet~\cite{chung2017out} and the distance between the landmarks of the mouth (\textbf{M-LMD})~\cite{chen2019hierarchical} for lip-sync evaluation. Furthermore, we measure the distance between the landmarks of the whole face (\textbf{F-LMD})~\cite{ji2022eamm} to evaluate the accuracy of the pose and facial expressions.

\textit{Visual quality.} 
We use \textbf{PSNR}, \textbf{SSIM}, and Frechet Inception Distance score (\textbf{FID}) to measure the image quality of synthesized video portraits.

\textit{Emotional accuracy.} 
To quantify the accuracy of the synthesized emotional video, We utilize the same emotion classifier mentioned in~\cite{gan2023efficient} for the MEAD dataset. For the CREMA-D dataset, we train the same classifier used in~\cite{eskimez2021speech} on the CREMA-D training set (total 76 identities).

\textit{Motion Diversity.} 
To evaluate the diversity and richness of generated head motions, following in ~\cite{zhang2023sadtalker}, we calculate the distance between various predicted 3-dimension head motion embeddings extracted by TokenHPE~\cite{zhang2023tokenhpe}, which can be defined as:
\begin{equation}
    {Div}=\frac{2}{B \times(B-1)} \sum_{i=1}^{B-1} \sum_{j=i+1}^B\left|\hat{m}_i-\hat{m}_j\right|_1,
\vspace{-0.1cm}
\end{equation}
where $\hat{m}_i$ and $\hat{m}_j$ are the $i$-th and $j$-th head motion embeddings in a batch $B$.

For the alignment of the audio and generated motions, we compute the beat align score proposed by~\cite{siyao2022bailando}. \textbf{BA} measures the average distance between each motion beat and its nearest corresponding audio beat:
\begin{equation}
    BA=\frac{1}{|B^m|} \sum_{b_j^m \in B^m} \exp \{-\frac{\min _{\forall b_j^a \in B^a}\left|b_i^m-b_j^a\right|^2}{2 \sigma^2}\},
\vspace{-0.1cm}
\end{equation}
where $B^m=\{b_i^m\}$ and $B^a=\{b_i^a\}$ denotes the motion beats and audio beats, respectively. Following~\cite{siyao2022bailando,liu2022beat}, we set the normalized parameter $\sigma$ as 3 in our experiment.

\begin{figure*}[t]
  \centering
  \includegraphics[width=0.98\textwidth]{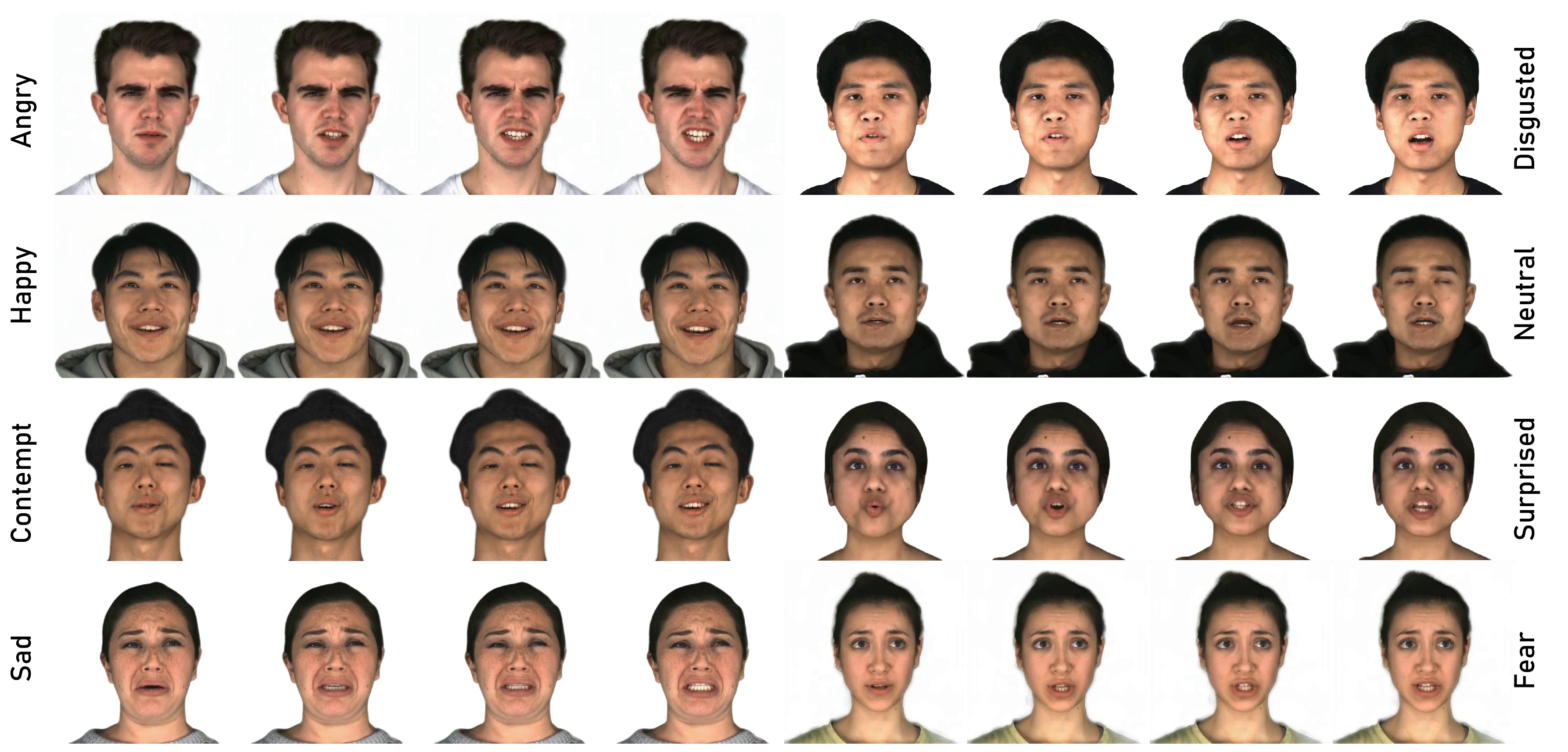}
  \vspace{-0.4cm}
  \caption{Additional emotional portraits generated by our GMTalker. Given the same input audio, we can generate high-fidelity and faithful emotional expressions according to the target emotion label. The identities are from the MEAD dataset.}
  \label{fig:multi_emo8}
\vspace{-0.4cm}
\end{figure*}

\begin{figure}[t]
  \centering
  \includegraphics[width=0.48\textwidth]{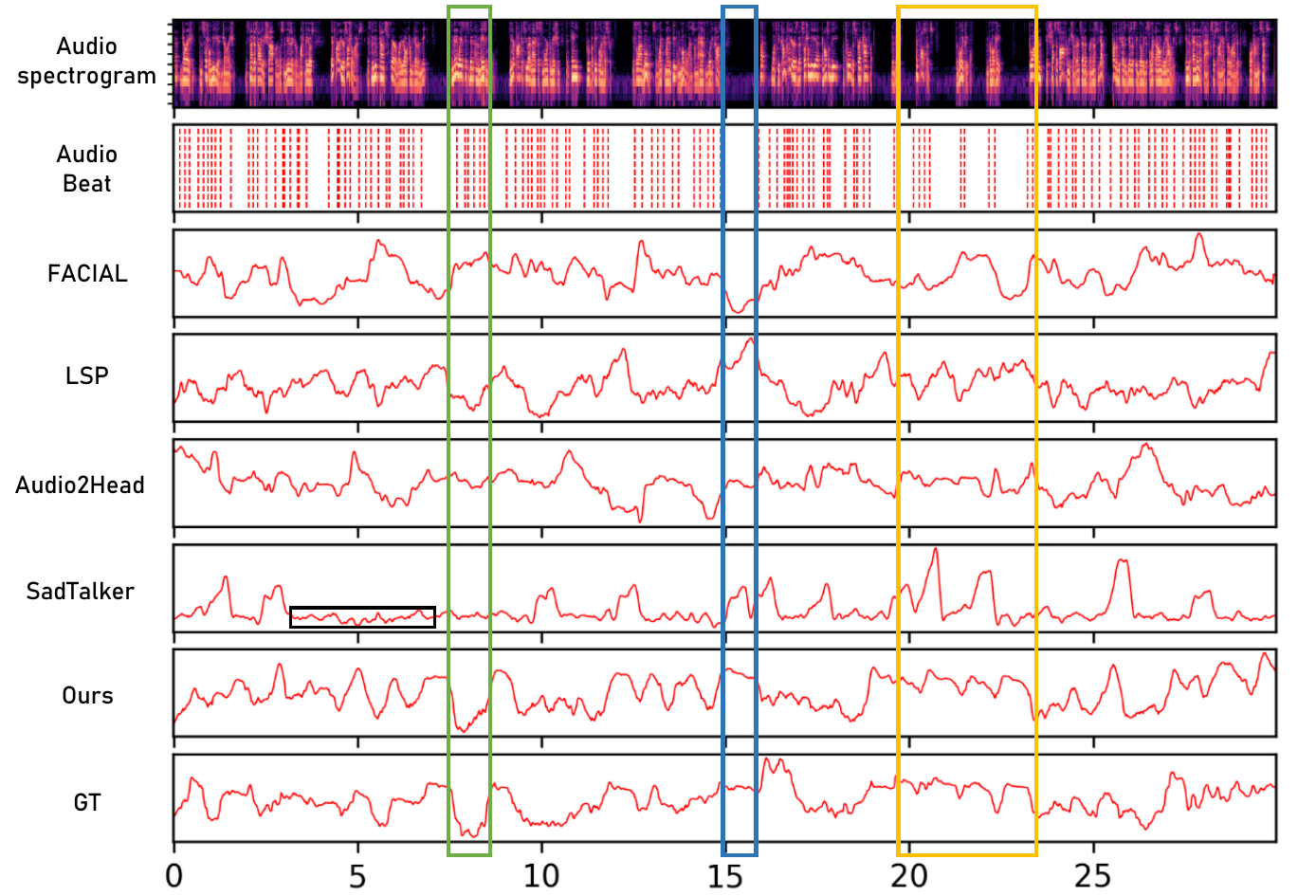}
  \vspace{-0.2cm}
  \caption{Comparison of head motion on the test video sample from LSP~\cite{lu2021live}. From top to bottom: the audio spectrogram, audio beat, FACIAL~\cite{zhang2021facial}, LSP~\cite{lu2021live}, Audio2Head~\cite{wang2021audio2head}, SadTalker~\cite{zhang2023sadtalker}, ours, and ground truth. The green box is the case of speaking, the blue box is the case of silence, and the yellow box is the case of speaking with pauses.}
  \label{fig:posecompare}
\vspace{-0.2cm}
\end{figure}

Besides, to measure the accuracy of predicted head motion, we compute the percentage of correctly predicted motion embeddings instead of keypoints, which can be calculated as:

\begin{equation}
    P C M=\frac{1}{T \times J} \sum_{t=1}^T \sum_{j=1}^J \boldsymbol{1}\left[\left|\hat{m}_t^j-m_t^j\right|_2<\tau\right],
\vspace{-0.2cm}
\end{equation}
where $J=3$ and $\hat{m}_t^j$, $m_t^j$ are the $j$-th dimension of predicted motion embeddings and the ground-truth motion embeddings at the $t$-th frame. We only calculate the successfully recalled motion embeddings against a specified threshold $\tau=1.0$. 

\subsection{Evaluation Metrics of Emotion Interpolation} \label{appendix:B_interp_metrics}
We further propose two metrics to validate the performance of emotion interpolation.

\textit{Emotion Perceptual Path Length.} We train a VGGNet~\cite{simonyan2014very} for emotion classification on the MEAD dataset to make the network more focused on emotion. Then, we calculate the mean perceptual distance between adjacent images in 17-frame sequences using the trained VGGNet. 

\textit{Emotion Perceptual Distance Variance.} Similarly, we compute the perceptual loss between adjacent images in 17-frame sequences and then calculate the variance of these distances in the sequence.

\section{More Experimental Results} 
\subsection{More Comparisions} \label{appendix:comparision}

Here, we first present the qualitative comparison on the CREMA-D dataset. As shown in Fig.~\ref{fig:cremadcomparsion}, our proposed method consistently generates high-fidelity emotional talking video portraits that accurately convey the intended expressions. In contrast to previous approaches, GMTalker reliably reproduces accurate expressions while preserving natural mouth shapes that synchronize with the input speech.

To comprehensively validate motion diversity and its correlation with audio, we utilize correlation map~\cite{wang2021audio2head} to compare our method with several motion-controllable methods.
We reduce the 3-dimensional head motion embeddings into one dimension by PCA following~\cite{wang2021audio2head}.
As depicted in Fig.~\ref{fig:posecompare}, existing methods struggle to generate realistic head movements consistent with rhythmic audio beats. 
In the case of silence audio (shown in the blue box), the head motion should remain static intuitively, yet their generated head motion often exhibits dynamic fluctuations.
Conversely, in the case of speaking audio (shown in the green box), the head motion should synchronize with the audio, but their generated head motion may remain static and lack rich changes.
In contrast, our head motions preserve the rhythm and synchronization with audio, and are much closer to the ground truth, as shown in the yellow box.

\begin{figure}[t]
  \centering
  \includegraphics[width=0.4\textwidth]{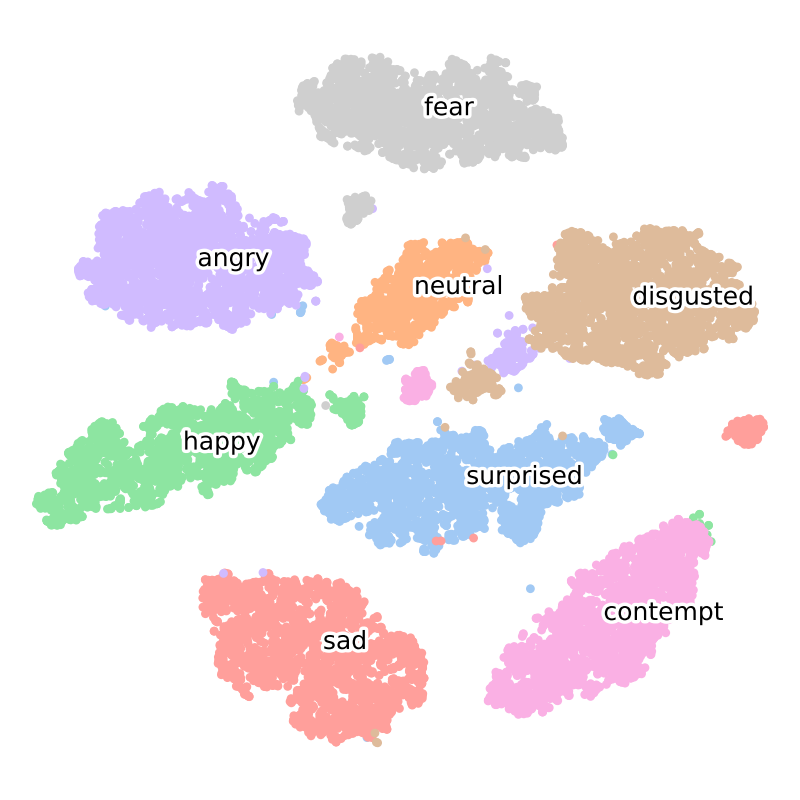}
  \vspace{-0.4cm}
  \caption{T-SNE visualization of Gaussian mixture latent space. Different colors indicate different emotion types. }
  \label{fig:tsne}
\vspace{-0.4cm}
\end{figure}

\input{table/user_study} 

\subsection{Disentanglement Analysis}
To better validate the disentangling of our GMEG, we feed the same input speech and different emotion labels into GMEG. As shown in Fig.~\ref{fig:multi_emo8}, the mouth movements in the generated video correspond to the speech, while the facial expression matches the target emotion label.
Moreover, to evaluate the disentanglement of various emotions in our Gaussian mixture latent space, we use t-SNE~\cite{van2008visualizing} to visualize the latent codes. Illustrated in Fig.~\ref{fig:tsne}, different colors represent sampled latent code with eight different emotion categories. By using the Gaussian mixture distribution, the samples sharing the same emotion are clustered together, while those with different emotions are distinctly separated. This indicates the contribution of the proposed GMEG in effectively disentangling various emotions from each other.

\begin{figure}[t]
  \centering
  \includegraphics[width=0.48\textwidth]{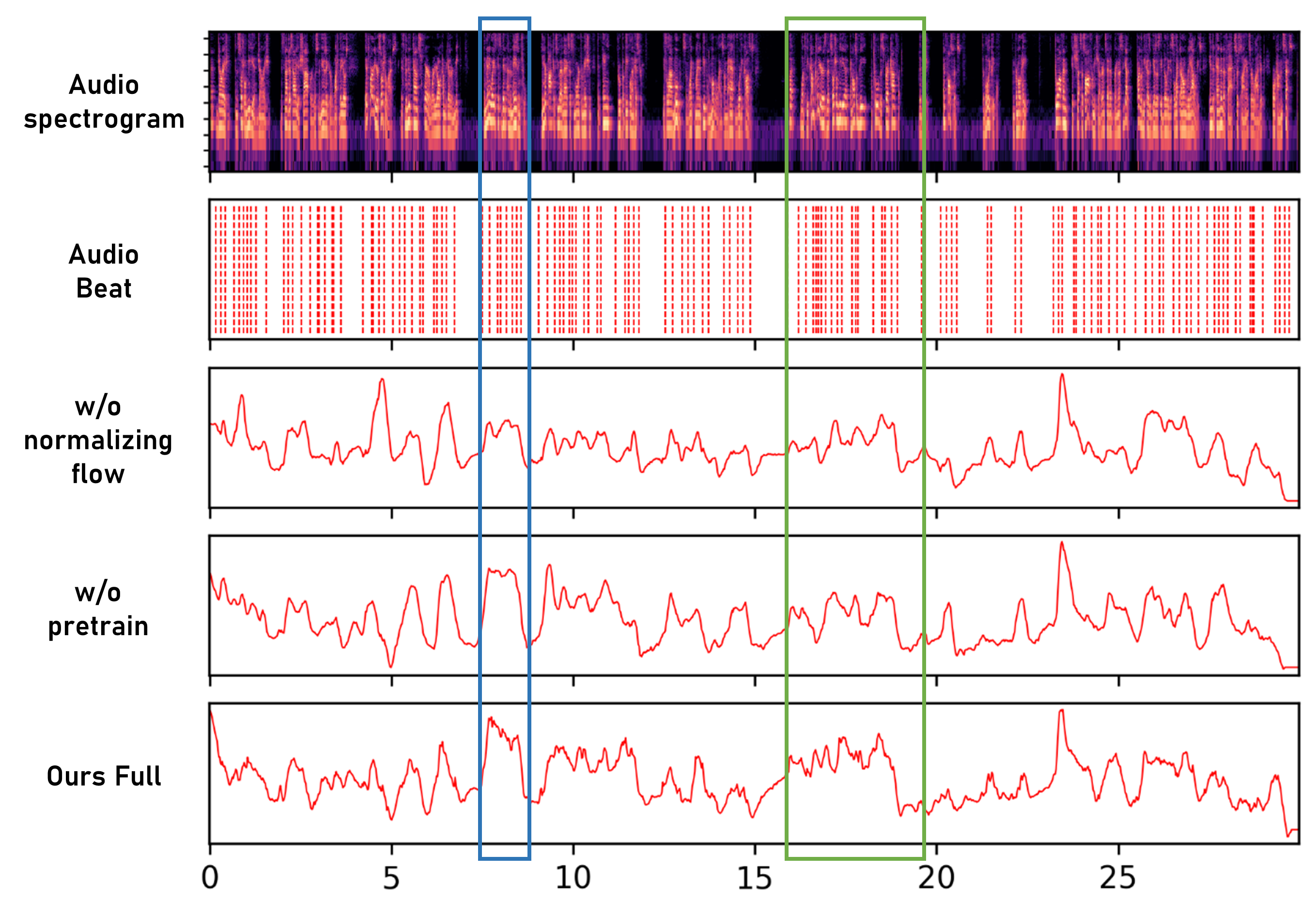}
  \vspace{-0.2cm}
  \caption{Qualitative results of ablation study for normalizing flow motion generator. From top to bottom: audio spectrogram, audio beat, the head motion beat w/o normalizing flow, w/o pre-train, and our full results. The boxes highlight the rhythm and synchronization of the head movements generated by our full model.}
  \label{fig:Ablation_pose}
\vspace{-0.2cm}
\end{figure}

\section{User Study} \label{appendix:D_use_study}
To further quantify the quality of generated video portraits, we conduct a user study to compare real data with generated ones from some representative methods: MakeItTalk~\cite{zhou2020makelttalk}, Styletalk~\cite{ma2023styletalk}, EVP~\cite{ji2021audio}, EAT~\cite{gan2023efficient}. 
We randomly select overall 24 audio clips (3 clips $\times$ 8 emotions) from the test set of MEAD to generate video samples for each method. The 14 recruited participants are required to evaluate the given video from three aspects “lip synchronization”, “video quality”, and “emotion accuracy” and choose the top two preferred videos for each of these aspects from all the methods presented. The results are shown in Table~\ref{tab:userstudy}. Our approach achieves the best scores for lip-sync, video quality, and emotion accuracy, indicating the expressiveness of our GMEG, NFMG, and emotion-guided head generator with EMN.

\section{More Ablation} \label{appendix:E_ablation}
To further demonstrate the effectiveness of the proposed NFMG, we conduct a perceptual study to evaluate the correlation between generated head motion and audio beat. 
As depicted in Fig.~\ref{fig:Ablation_pose}, the head movements generated by our full model exhibit greater rhythm and synchronization with the audio beat, while also showing increased diversity.

%% file: table/user_study.tex
\begin{table}[t]
\centering
\caption{User Study on MEAD datasets. The table displays the percentage of participants' preferences for each method in terms of each aspect.}
\vspace{-0.2cm}
\setlength\tabcolsep{2.0pt} 
\begin{tabular}{lcccccc}
\toprule
Method & MakeItTalk & Styletalk & EVP & EAT & GMTalker & GT \\
\midrule
Visual Quality & 4.19 & 5.18 & 10.35 & 6.99 & \textbf{35.38} & 37.91 \\
Audio-visual Sync& 7.92& 6.13 & 5.08 & 12.41 & \textbf{29.45} & 39.01 \\
Emotion Accuracy & 4.46 & 5.94 & 4.75& 10.55&\textbf{34.77} & 39.52 \\

\bottomrule
\end{tabular}
\vspace{-0.4cm}
\label{tab:userstudy}
\end{table}